\pdfoutput=1
\RequirePackage{fix-cm}
\documentclass[twocolumn]{svjour3}          % twocolumn
\smartqed  % flush right qed marks, e.g. at end of proof

\usepackage{graphicx}

\usepackage{url}
\usepackage{xcolor}
\usepackage{hyperref}
\hypersetup{
     colorlinks=true,
     linkcolor=orange,
     filecolor=orange,
     citecolor=orange,      
     urlcolor=orange,
     }
\usepackage{color, colortbl}
\usepackage{amsfonts}
\usepackage{amsmath}
\usepackage{appendix}
\usepackage{algorithm, setspace}
\usepackage{algpseudocode}
\usepackage{csquotes}
\usepackage{amsfonts}
\usepackage{booktabs}
\usepackage{siunitx}
\usepackage{breqn}
\usepackage[normalem]{ulem}
\usepackage{tikz}
\usetikzlibrary{tikzmark}

\newcommand{\p}[1]{\medskip \noindent \textbf{{#1}.}}
\newcommand{\eq}[1]{Equation~(\ref{eq:#1})}
\newcommand{\fig}[1]{Figure~\ref{fig:#1}}
\newcommand{\alg}[1]{Algorithm~\ref{alg:#1}}

\DeclareMathOperator*{\argmax}{arg\,max}

\journalname{Autonomous Robots}

\DeclareUnicodeCharacter{2212}{-}

\begin{document}

\title{VIEW: Visual Imitation Learning with Waypoints}
% \subtitle{Do you have a subtitle?\\ If so, write it here}


    \author{Ananth Jonnavittula \and Sagar Parekh \and Dylan P. Losey}

% \authorrunning{Short form of author list} % if too long for running head

\institute{A. Jonnavittula \at
              Mechanical Engineering Department, Virginia Tech \\
              \email{ananth@vt.edu}           %  \\
           \and
           S. Parekh \at
              Mechanical Engineering Department, Virginia Tech \\
              \email{sagarp@vt.edu}           %  \\
           \and
           D. Losey \at
            Mechanical Engineering Department, Virginia Tech \\
            \email{losey@vt.edu}
}

\maketitle

\begin{abstract}
Robots can use Visual Imitation Learning (VIL) to learn manipulation tasks from video demonstrations. However, translating visual observations into actionable robot policies is challenging due to the high-dimensional nature of video data. This challenge is further exacerbated by the morphological differences between humans and robots, especially when the video demonstrations feature humans performing tasks. To address these problems we introduce \textbf{V}isual \textbf{I}mitation l\textbf{E}arning with \textbf{W}aypoints (VIEW), an algorithm that significantly enhances the sample efficiency of human-to-robot VIL. VIEW achieves this efficiency using a multi-pronged approach: extracting a condensed prior trajectory that captures the demonstrator's intent, employing an agent-agnostic reward function for feedback on the robot's actions, and utilizing an exploration algorithm that efficiently samples around waypoints in the extracted trajectory. VIEW also segments the human trajectory into grasp and task phases to further accelerate learning efficiency. Through comprehensive simulations and real-world experiments, VIEW demonstrates improved performance compared to current state-of-the-art VIL methods. VIEW enables robots to learn manipulation tasks involving multiple objects from arbitrarily long video demonstrations. Additionally, it can learn standard manipulation tasks such as pushing or moving objects from a single video demonstration in under 30 minutes, with fewer than 20 real-world rollouts. Code and videos here: \url{https://collab.me.vt.edu/view/}

\end{abstract}


\section{Introduction}

Imagine teaching a person to pick up a cup placed on a table. The quickest method is often to physically demonstrate this task. Through physical demonstration, the observer can discern which object to pick up and how to manipulate that object. Humans efficiently learn everyday tasks in this way, including moving items, making tea, or stirring the contents of a pan.

Teaching robots these same tasks, however, proves to be more cumbersome. Typically, robots employ either Imitation Learning (IL) or Reinforcement Learning (RL) methods. IL generally requires many demonstrations from humans to obtain effective policies \cite{fang2019survey,hussein2017imitation}. During this process, the human teacher often needs to kinesthetically guide or teleoperate the robot to show it exactly what actions it should take. On the other hand, RL methods require a substantial number of rollouts for robots to perform even simple tasks \cite{kober2013reinforcement,morales2021survey}. Additionally, defining appropriate reward functions for reinforcement learning poses a challenge, particularly in unstructured environments \cite{dulac2021challenges}.

\begin{figure*}
    \centering
    \includegraphics[width=2\columnwidth]{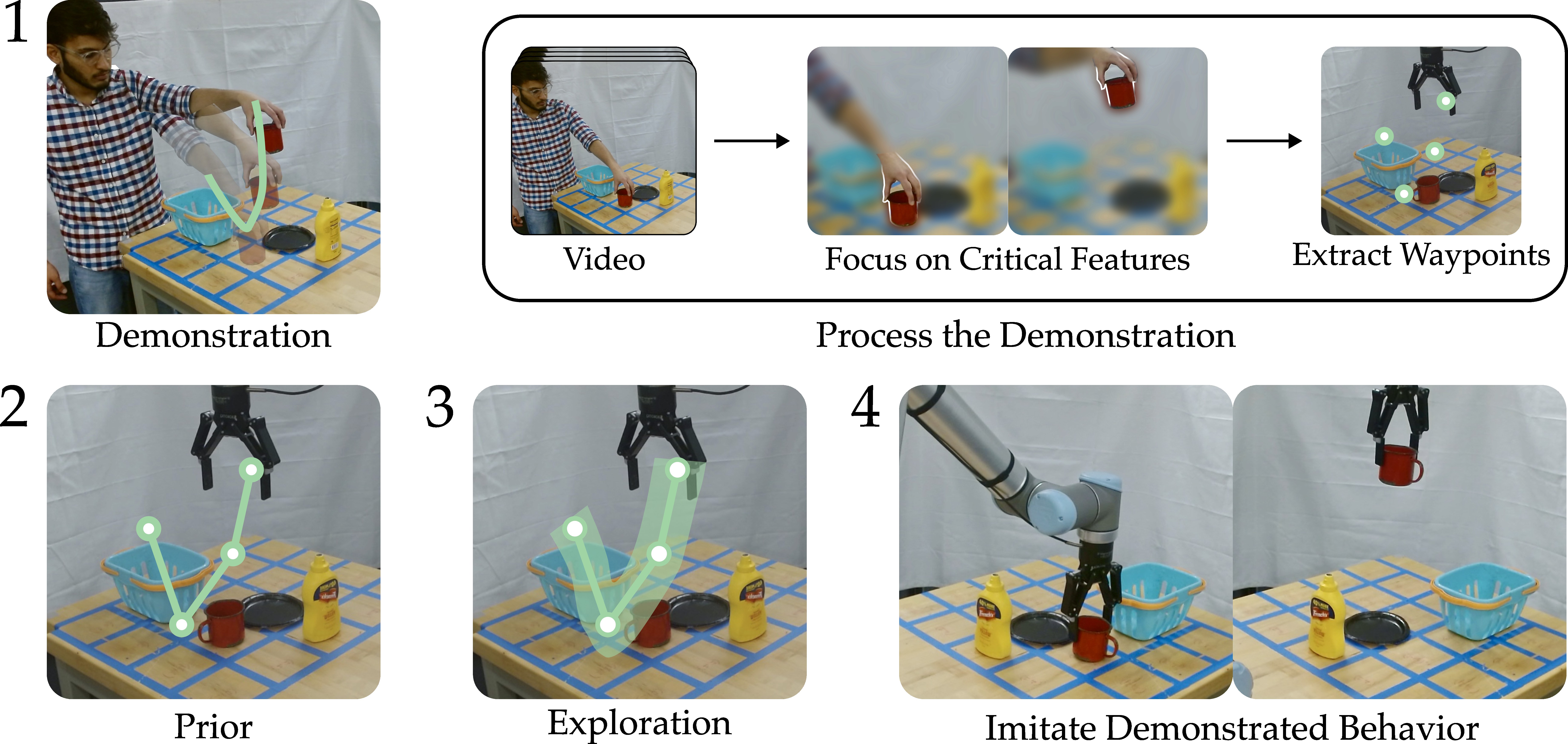}
    \caption{Robot learning from visual demonstration. 1) A human demonstrates the task directly in the environment: here we use the example of picking up a cup. Under our proposed approach, the robot processes a single video of that demonstration to selectively focus on important features such as the human hand and the manipulated object. From these trajectories the robot obtains \textit{waypoints} that capture the critical parts of the task (e.g., grasping the cup). 2) These extracted waypoints serve as a prior for the correct robot trajectory. 3) In practice, simply executing this prior rarely leads to task success due in part to the morphological differences between human and robot (in this case, the robot misses the cup entirely). Therefore, the robot must explore in a region around the initial waypoints to iteratively improve its trajectory. 4) After repetitively interacting with the environment, the robot learns to successfully imitate the behavior demonstrated in the human video.}
    \label{fig:front}
\end{figure*}

In this paper we therefore study how robots can learn tasks by \textit{watching} humans. The human provides a demonstration directly in the environment (e.g., physically picking up a cup), and the robot collects an RGB-D video of the human's demonstration. Our objective is for the robot to leverage this single video to learn the task and correctly manipulate the same object. The primary issue here lies in the overload of information conveyed by video demonstrations. Each video is comprised of thousands of image frames, and each frame contains numerous pixels. These raw pixel values --- when seen in isolation --- are not sufficient for the robot to determine what actions it should take (i.e., how to move the robot arm). Consequently, robots that learn from video demonstrations must extract pertinent information from a large amount of data.

To address this fundamental problem, our hypothesis is that robots do not need to reason over \textit{all} the video data. Consider our motivating example of picking up a cup: when humans learn by watching other humans, we do not focus on environmental clutter or extraneous details. Instead, we just need to see where the human grabs the cup and how they carry it. At a high level, we can think about these critical parts of the task as \textit{waypoints}: the cup's initial position, the human's hand configuration when grasping, key frames along the cup's motion, and where the human finally places the cup. Robots that can learn these waypoints from the human's video demonstration will be able to perform the overall task without having to reason over every single aspect of every video frame.

We leverage this hypothesis to develop VIEW: Visual Imitation lEarning with Waypoints (see \fig{front}).
Our approach starts with a video of the human performing their desired manipulation task.
We then process that video to get an initial trajectory (e.g., a best guess) of how the robot should complete the same task.
To obtain this initial guess we extract the human's hand trajectory, and then autonomously identify the critical waypoints along that trajectory in visual and Cartesian spaces.
In an idealized scenario, the robot could directly execute this initial trajectory and complete the task.
However, the initial trajectory almost always fails because of (a) the morphological differences between human demonstrator and robot learner and (b) the sensor noise in the initial RGB-D video.
Returning to our cup example, we often find that --- even though the video shows the human picking up the cup --- the robot's extracted trajectory misses that cup entirely.

Accordingly, the second part of VIEW focuses on iteratively improving the robot's prior and correctly completing the task. We develop sampling strategies so that the robot can intelligently explore around its waypoints.
This includes waypoints where the robot needs to grasp an object (e.g., pick up the cup) and waypoints where the robot is manipulating that object (e.g., carrying the cup to a goal location). Again, we rely on our hypothesis: instead of reasoning about every aspect of the video, we focus on the object's location in the waypoint frames. This leads to an iterative learning process where the robot corrects its initial trajectory and eventually completes the original task shown in the video. Overall, VIEW enables robot arms to efficiently learn manipulation tasks such as picking, pushing, or moving objects, requiring fewer than $20$ real-world trials and less than $30$ minutes from demonstration collection to successful task execution. Additionally, our method enables robots to learn from long horizon videos that involve a combination of tasks --- such as moving a cup and dispensing tea into it, or placing multiple objects in a pan --- using no additional information besides the initial human demonstration video.

Overall, we make the following contributions:

\p{Condensed prior extraction} We present a new approach for distilling a prior from video demonstrations that accurately reflects the human demonstrator's intent. We achieve this by extracting a concise set of waypoints that capture the human hand trajectory and its interaction with objects.

\p{Agent agnostic rewards} For sample efficient exploration, the robot requires effective feedback when exploring around the waypoints in the extracted prior. To provide this feedback, we propose an agent-agnostic reward function. Our reward model only focuses on the critical components of the task --- i.e., movement of the object --- regardless of which agent performs the task.

\p{Sample efficient exploration} Given morphological disparities and noise introduced during prior extraction, directly replicating the human trajectory is impractical. We introduce an algorithm that segments the prior into \textit{grasping} and \textit{task} phases, sequentially focusing on locating the object and replicating its movement.

\p{Few shot adaptation} While our approach can bridge the embodiment gap between human and robot through efficient exploration around the human prior, each new task requires starting from scratch. However, the robot gains valuable insights into the morphological differences and camera noise with each solved task. By integrating a residual learner that leverages this insight with our prior extraction, we demonstrate that the robot can achieve few-shot learning on new tasks.

\p{Comparing VIEW to baselines} We conduct a comparative analysis of our method against existing state-of-the-art approaches in visual imitation learning. Additionally, an ablation study is performed in a simulated environment to underscore the significance of each component within our framework. These comparative analyses and ablation studies collectively demonstrate our method's efficacy in enabling robots to quickly imitate a wide range of tasks based on video demonstrations.
\section{Related Work}

We study how robots can efficiently learn to replicate a task based on a single video demonstration. Our approach builds upon existing learning from demonstration methods, particularly those that use videos, waypoints, and human activity recognition.

\p{Learning from demonstrations (LfD)} LfD is a general learning framework  \cite{pomerleau1991efficient,schaal1996learning} that is used across domains such as autonomous driving \cite{pan2020imitation,chen2019deep}, robotics \cite{jonnavittula2021know,jonnavittula2021learning,jonnavittula2022sari,mehta2023unified,ratliff2007imitation}, and video games \cite{amiranashvili2020scaling,schafer2023visual,scheller2020sample}. In robotics, LfD has been employed to learn from teleoperated expert demonstrations \cite{jonnavittula2022sari,ross2011reduction,kelly2019hg,menda2019ensembledagger}, extended to include imperfect demonstrations  \cite{jonnavittula2021know,brown2020better,brown2019extrapolating}, and combined with other modalities such as preferences \cite{mehta2023unified,taranovic2022adversarial,habibian2022here} or language \cite{shi2024yell,lynch2020language,liu2024ok}. A significant aspect of LfD in robotics involves the sourcing of demonstrations, predominantly obtained from human actions within the agent's environment. For example, in autonomous driving, demonstrations encompass steering controls similar to those the agent uses \cite{pan2020imitation}, while in robotic manipulation, demonstrations are acquired via direct teleoperation \cite{jonnavittula2021learning} or kinesthetic teaching \cite{mehta2023unified}. This reliance on human provided demonstrations presents certain challenges, especially in robotics. Humans primarily use their hands for manipulation, whereas robots utilize end-effectors with distinct morphologies. This fundamental discrepancy limits the feasibility and diversity of the training data collected for robot learning.

To address the morphological disparities between humans and robots, some researchers have advocated for the use of tools such as reacher-grabbers that resemble grippers commonly employed in robotics \cite{pari2021surprising,shafiullah2023bringing,song2020grasping,young2021visual}, or utilizing camera angles that reduce the effects of hand to gripper morphology \cite{kim2023giving,duan2023ar2}. These tools facilitate the recording of demonstrations that can be more easily translated into actionable robot policies, without the need for teleoperation. While these approaches have proven effective, they do not ameliorate the underlying limitation: the demonstrations are inherently restricted in scope due to the specialized interface. In response to this challenge, there has been a shift towards compiling extensive robot demonstration datasets, like Open-X \cite{padalkar2023open}, aiming to establish a foundational resource akin to ImageNet for robot learning. But these datasets overlook the vast reservoir of already existing human video demonstrations, which could significantly expedite the learning process for robots. VIEW builds upon prior LfD works by learning from human demonstrations. However, VIEW focuses on learning directly from human videos, and does not rely on kinesthetic demonstrations, teleoperated inputs, or intermediary tools.

\p{Learning from video demonstrations} There has been a parallel research focus on teaching robots with videos of robots performing the desired task \cite{cetin2021domain,rafailov2021visual,wen2021keyframe,wen2022you}. In these methods a human teleoperates the robot in the video demonstration (i.e.,  \textit{the video demonstration is of the robot completing the task}), and the robot learns to imitate the resulting RGB-D video. These methods tackle the challenges that arise from a lack of explicit action information \cite{cetin2021domain,wen2021keyframe,wen2022you}. A significant aspect of this research is the reduction of data complexity through an emphasis on keyframes \cite{wen2021keyframe,wen2022you}, aiming to simplify robot learning by concentrating on achieving these specific frames. Nevertheless, these methods necessitate a large number of trials with the robot \cite{cetin2021domain}, and do not solve the key problem of obtaining generalized video demonstrations from humans or other agents.

Alternatively, some researchers have explored learning from \textit{human video demonstrations} directly, with considerable effort dedicated to transforming human videos into a format applicable to the robot's domain \cite{smith2019avid,li2021meta,xiong2021learning,sharma2019third}. To facilitate this transformation, these methods utilize cycle consistency networks \cite{isola2017image} to translate human videos into equivalent robot videos \cite{xiong2021learning,sharma2019third}. Once videos are translated, they extract key points from the videos, which serve as the basis for learning \cite{xiong2021learning,liu2018imitation}. However, a significant drawback of this approach is the necessity for a vast collection of videos, showcasing both humans and robots performing tasks. This requirement poses a substantial limitation, adversely affecting the scalability of such methods.

Several approaches closely align with our method, focusing on \textit{human-to-robot imitation} learning \cite{lee2022learning,jain2024vid2robot,jin2020geometric,sieb2020graph,alakuijala2023learning,sermanet2017unsupervised,chane2023learning,Patel2022,shaw2024learning,whirl}. These methods extract meaningful representations of a task from videos \cite{chane2023learning} or use neural networks to learn reward functions from the videos to facilitate reinforcement learning \cite{sermanet2017unsupervised,alakuijala2023learning}. Despite the promise shown by many of these methods, they share a common challenge: the necessity for a substantial number of robot rollouts in real-world scenarios to learn tasks effectively. One particularly similar approach here is WHIRL \cite{whirl}, which mirrors our method but employs a variational autoencoder-based exploration exploitation strategy. As we will show, WHIRL requires a large number of rollouts to converge, and struggles to scale for long horizon tasks. Additionally, this approach also requires video inpainting \cite{lee2019copy}, where the human must be removed from each frame in the demonstration video, and the robot must be removed from each rollout frame before reward computation. Given that inpainting is highly GPU-intensive, this process can require powerful GPUs and lead to substantial wait times between rollouts. Our proposed method, VIEW, addresses these challenges by segmenting and sequentially solving the task, reducing computational demands. In our experiments, we will directly compare VIEW and WHIRL in terms of task success and training time to highlight these improvements.

\p{Waypoint-based learning} While the methods discussed so far primarily focus on learning action policies directly from demonstrations, a growing trend in robotics is a shift towards teaching robots to reach designated waypoints. This approach is gaining traction, particularly because it aligns well with the use of separate planning and control algorithms, allowing low-level controllers to reach goals set by high-level planners. This concept has seen application in reinforcement learning for tasks such as object pick-and-place and door opening \cite{mehta2024waypoint}. However, the success of these algorithms hinges on the creation of meticulous reward functions to guide learning. In the domain of imitation learning, it has facilitated the learning of intricate tasks, such as operating a coffee machine \cite{shi2023waypoint}. Nevertheless, similar to methods discussed on learning from video demonstrations, Shi \textit{et al.} \cite{shi2023waypoint} learn from a video demonstration of the robot performing the task. This distinction is crucial, as robot demonstrations bypass the morphological differences encountered when learning from human videos. Our approach to prior compression bears similarities to these waypoint-focused methods. However, it differs in that VIEW learns directly from a single human video, where the human physically performs the task without a robot.

% \p{Human activity recognition} Many of the methods discussed above rely on human intent and activity recognition for enabling robots to understand object affordance. Within the realm of robot manipulation, numerous studies have proposed methods for discerning human actions and activities from video footage \cite{eze2024learning,luo2023learning,koppula2013learning,koppula2015anticipating}. Some works extend human posture tracking to identify fine-grained activities within limited spatial contexts \cite{ma2018region}. However, merely detecting human presence is insufficient to learn from human videos. To bridge this gap, using annotated video datasets such as SomethingSomething \cite{goyal2017something}, YouCook \cite{das2013thousand}, ActivityNet \cite{caba2015activitynet}, or the 100 Days of Hands (100DOH) \cite{shan2020understanding} becomes instrumental. The 100DOH dataset is particularly valuable due to its detailed object interaction annotations. Building upon prior works, our approach utilizes the 100DOH framework to extract crucial data on hand positioning and interactions with objects.

\p{Human activity recognition} Many of the methods discussed above rely on human intent and activity recognition for enabling robots to understand object affordance. Within the realm of robot manipulation, numerous studies have proposed methods that use annotated video datasets such as SomethingSomething \cite{goyal2017something}, YouCook \cite{das2013thousand}, ActivityNet \cite{caba2015activitynet}, or the 100 Days of Hands (100DOH) \cite{shan2020understanding}. The 100DOH dataset is particularly valuable due to its detailed object interaction annotations. Building upon prior works, our approach utilizes the 100DOH framework to extract crucial data on hand positioning and interactions with objects.

\smallskip

In contrast to the many of the methods discussed here, VIEW distinguishes itself by focusing on sample-efficient learning directly from human videos. Our approach aims to teach robots manipulation tasks --- such as picking or moving objects --- with minimal human supervision. The only interaction required from the human is providing a single video demonstration.
\section{Problem Statement}
\label{sec:problem}

We consider single object manipulation tasks in unstructured environments. First a human teacher physically demonstrates their desired task within the robot's workspace. During this demonstration the robot is moved out of the way (i.e., the human does not interact with the robot) and the robot records the human's behavior with a stationary RGB-D camera. 
After the demonstration is complete the video is provided to the robot, and the robot must learn how to replicate the same task based on this single video. We highlight that the human and robot have morphological differences --- e.g., the human's hand is different from the robot's gripper --- and so the way the human performed the task may not transfer directly to the robot arm.

\begin{figure*}[t]
    \centering
    \includegraphics[width=2.0\columnwidth]{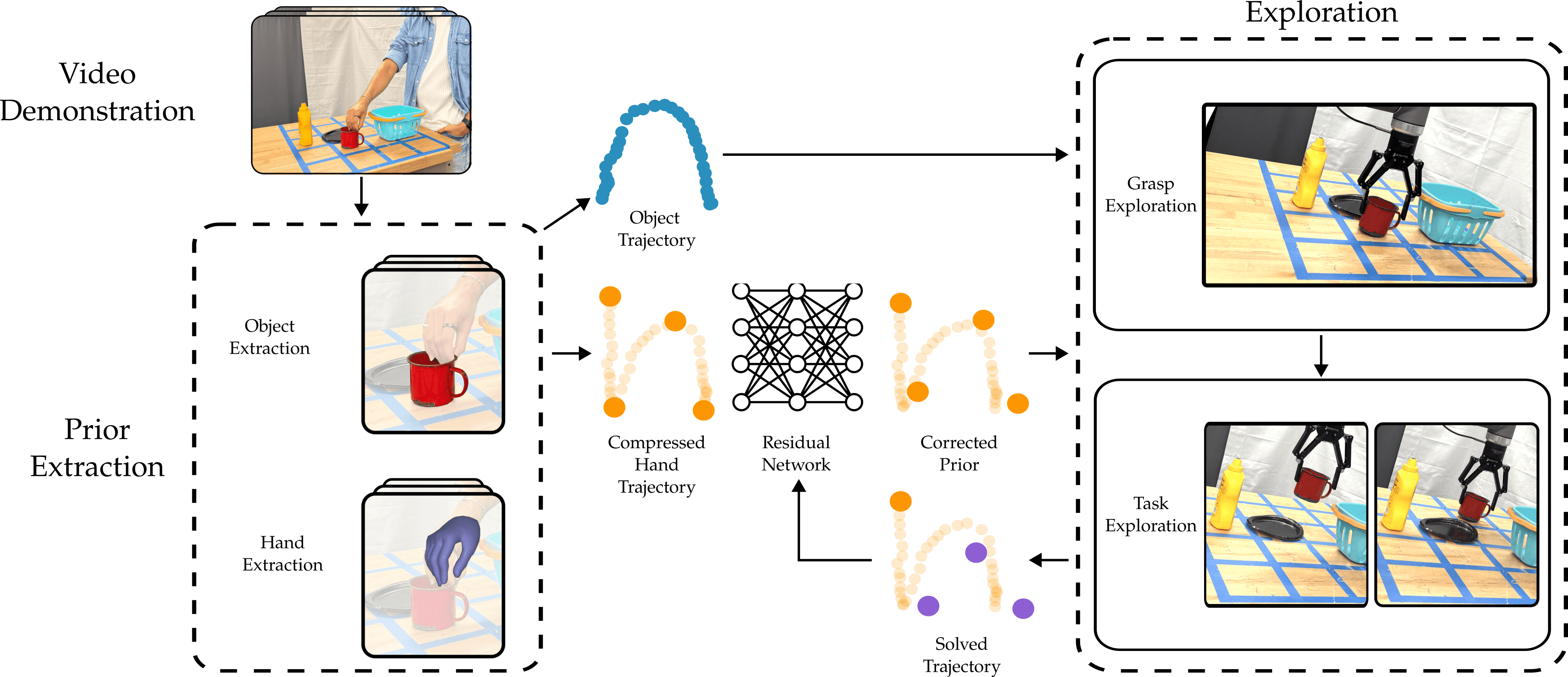}    
    \caption{Outline of VIEW, our proposed method for human-to-robot visual imitation learning. (Top Left) VIEW begins with a single video demonstration of a task. (Bottom Left) From this video we extract the object of interest, its trajectory, and the human's human trajectory. (Middle) We then perform compression to obtain a trajectory prior --- a sequence of waypoints the robot arm should interpolate between to complete the task. Unfortunately, this initial trajectory is often imprecise due to the differences between human hands and robot grippers, as well as noise in the extraction process. We therefore refine the prior using a residual network, which is trained on previous tasks to de-noises the current data. (Right) The de-noised trajectory is then segmented into two phases: grasp exploration and task exploration. (Top Right) During grasp exploration, the robot determines how to pick up the object by modifying the pick point in its trajectory. (Bottom Right) Following a successful grasp, the robot proceeds to task exploration, where is simultaneously corrects the remaining waypoints of the trajectory. After completing exploration, the robot synthesizes a complete trajectory. (Middle) This solved trajectory, alongside the prior trajectory, is used to further train the residual network, thus enhancing the performance of our method in future tasks.}
    \label{fig:method}
\end{figure*}

\p{Environment} We formulate the robot's environment as a Markov Decision Process without rewards: $\mathcal{M} = \langle \mathcal{S}, \mathcal{A}, \mathcal{T} \rangle$. The robot's end-effector position in Cartesian space is its state $s$, and the robot's workspace becomes its state space $\mathcal{S}$. In every state the robot can take an action $a \in \mathcal{A}$ which is the end-effector velocity. This action moves the robot to a new state $s'$ based on the environment transition probability $\mathcal{T}(s'\mid s, a)$. We assume the environment remains unchanged between the human demonstration and the robot's learning activity; that is, all objects maintain their positions. Additionally, we only consider scenarios where the environment is captured from a fixed camera perspective.

\p{Video Demonstration} The robot learns from a single video demonstration ($V_i$) of a human performing the task ($\tau_i$). This video is captured using a stationary RGB-D camera that is set up such that the human and the object they manipulate are visible at all times. The human does not interact with the robot beyond providing this video. Although the human only provides one video for task $\tau_i$, they may provide demonstrations for multiple tasks: e.g., the robot could receive a set of $n$ videos $V_1, \ldots, V_n$ for $n$ different tasks $\tau_1, \ldots, \tau_n$. The robot's objective is to map each video into a trajectory that completes the demonstrated task.

\p{Tasks} Our focus is primarily on tasks that involve manipulating a single object at a time. We consider tasks such as picking, pushing, or moving an object within the robot's workspace. While multi-object tasks are allowed, they must be decomposed into multiple sequential single-object manipulation tasks, similar to those mentioned above.
\section{VIEW} \label{sec:view}

Our approach to imitation learning from video demonstrations relies on our intuition that efficient learning requires focusing on critical waypoints. In this section we discuss our approach that consists of three main parts: first, extracting which object to pay attention to, how this object moves throughout the task. Second, designing a robust reward signal that compares the robot's behavior to the human's behavior. Third, exploring around the extracted waypoints in a sample-efficient manner. We refer to our method as \textbf{VIEW:} \textbf{V}isual \textbf{I}mitation l\textbf{E}arning with \textbf{W}aypoints. Refer to \fig{method} for an overview.

\subsection{Prior Extraction} \label{sec:view_prior}

% VIEW relies on three crucial pieces of information extracted from the human's video demonstration: identification of the manipulated object, understanding of the object's movement within the video, and the human's hand movements during the task. Consequently, the prior extraction process can be divided into two main subparts: one concerning the object and its motion in the video, and the other focusing on the human's interactions with that object. Therefore, we separate our prior extraction approach into two distinct components: Hand Trajectory Extraction and Object Trajectory Extraction. A summary of our overall prior extraction method can be found in \fig{prior}.

\begin{figure*}[t]
    \centering
    \includegraphics[width=2\columnwidth]{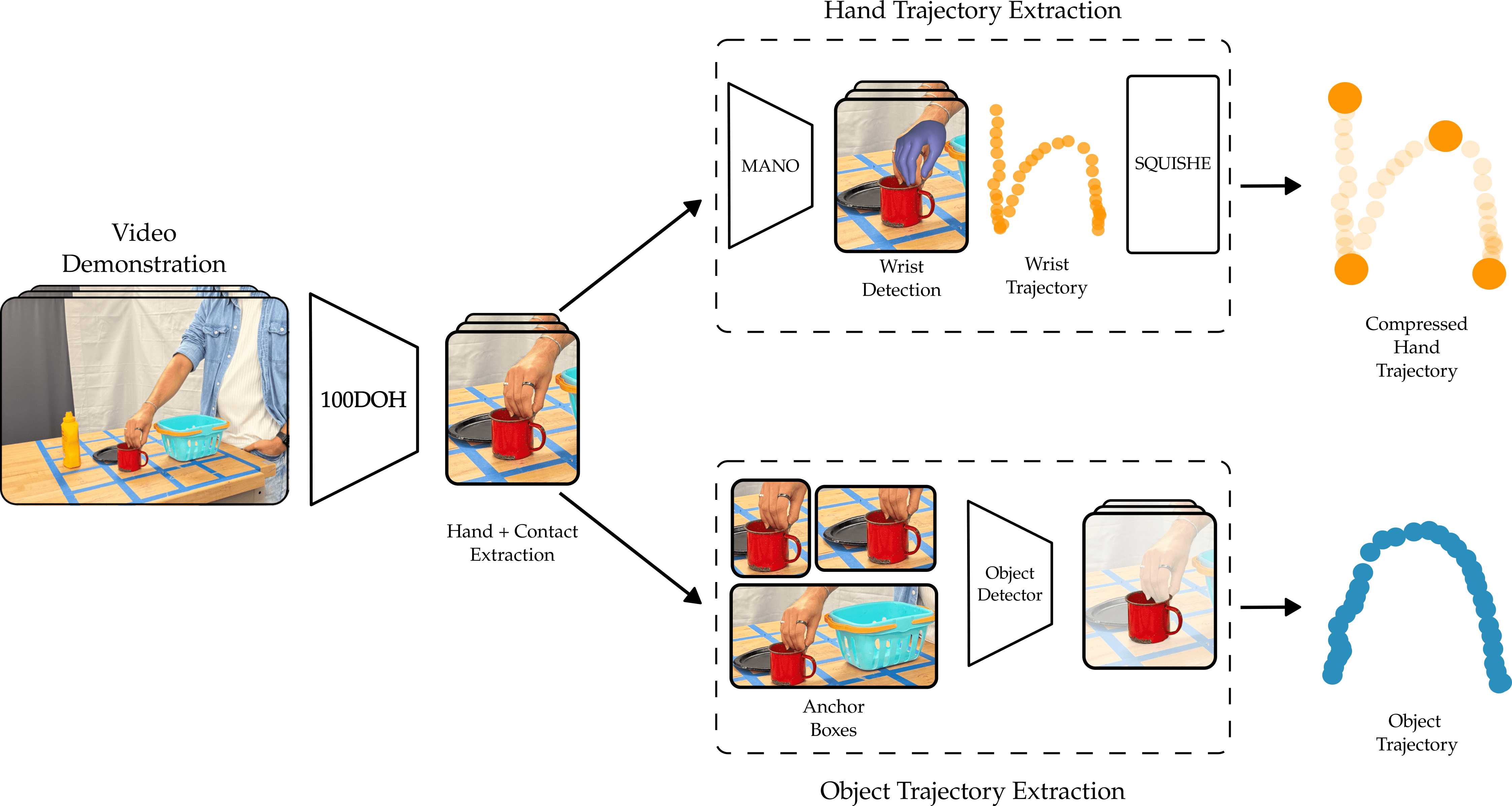}    
    \caption{An overview of our prior extraction method (Bottom Left in \fig{method}). Utilizing the 100 Days of Hands ($100$DOH) detector \cite{shan2020understanding}, we first identify the location of the hand and if it is in contact with any objects present in the frame. We then refine the human's hand trajectory using the MANO model \cite{romero2022embodied} to capture wrist movements. Subsequently, to eliminate redundancy, we apply the SQUISHE algorithm \cite{muckell2014compression}. This produces an initial trajectory with key waypoints that the robot should interpolate between. To pinpoint the object of interest amidst potential clutter, we analyze frames where hand-object contact occurs, creating anchor boxes that --- in conjunction with an object detector --- reveal the object the human interacts with most frequently. This identification enables us to construct an accurate object trajectory from the human's video.}
    \label{fig:prior}
\end{figure*}

Since VIEW relies on information about both hand and object movements, we divide the extraction process into two components: Hand Trajectory Extraction and Object Trajectory Extraction. A summary of our overall prior extraction method can be found in \fig{prior}.

\p{Hand Trajectory Extraction} 
Prior methods have extensively addressed the extraction of hand trajectories from video demonstrations, often leveraging open-source neural networks for this purpose \cite{whirl,wang2020rgb2hands,zhang2020mediapipe}. In our approach (see \fig{prior}), we analyze each frame ($v_t$) in a video ($V$) using the 100 Days of Hands (100DOH) model \cite{shan2020understanding}. This model helps us identify the hand’s location and whether it is interacting with any objects via bounding box coordinates ($b^h_{x_t}, b^h_{y_t}$) and contact information ($c_t$). A bounding box alone can be ambiguous with respect to hand orientation and direction. To resolve this ambiguity, we further refine our coordinates with the MANO hand model \cite{romero2022embodied} to pinpoint the human's wrist position ($p^h_{x_t}, p^h_{y_t}$). We convert the $2$D image coordinates into $3$D world coordinates ($x^h_{t},y^h_{t},z^h_{t}$) using depth information from the camera. We use \textit{waypoint} to refer to the $3$D world coordinates of the human hand or the object.

So far our methodology bears a strong resemblance to that used in WHIRL \cite{whirl}. However, WHIRL overlooks a significant issue: the abundance of points along the extracted trajectory. For instance, a mere ten-second demonstration recorded at $60$ frames per second yields a total of $600$ frames. While this amount of visual and trajectory data appears substantial, upon closer inspection, much of it is redundant. Returning to our example of picking up a cup, the video contains crucial waypoints such as the initial hand position and the cup's grasp position, but it also includes several redundant frames that interpolate between these key waypoints. This principle extends to various manipulation tasks; for instance, stirring a pan necessitates waypoints depicting the start location, spatula grasp, stirring locations, and final place location. All other intermediate points can be discarded.

To take advantage of this redundancy, we apply a trajectory compression algorithm called Spatial Quality Simplification Heuristic - Extended (SQUISHE) \cite{muckell2014compression}. SQUISHE enables users to prioritize key waypoints by setting a maximum allowable error or target count, selectively removing points that do not affect the  trajectory shape. To achieve this, it minimizes synchronized Euclidean distance by calculating the difference between each waypoint and its interpolated position, pruning points that fall within an acceptable error range. For instance, if the movement between the hand’s initial position and the cup grasp can be interpolated, SQUISHE removes intermediate waypoints. By condensing the trajectory in this way, we retain only significant trajectory changes, such as shifts in hand direction or contact with objects, often reducing the length from over $300$ points to just $3$ or $4$.

\p{Object Trajectory Extraction} Thus far, we have focused on extracting a concise prior trajectory from the human's hand movements. However, merely mimicking human actions is not sufficient for the robot to solve the task. In reality, the critical aspect of these demonstrations lies in understanding how the human is interacting with and moving objects. Revisiting our cup example, the focus should not be on repeating the human's hand movements; instead, the robot must learn to move the cup to the correct location.

% To extract the object trajectory we start by identifying which object the human is manipulating. Here we capitalize on the human's hand interactions in the video demonstration. From the hand trajectory extraction, we know that the 100DOH model can indicate when the human's hand interacts with objects in a frame ($c_t$). Building on this insight, we initiate a process akin to anchor boxes for region proposal networks \cite{ren2015faster} in image detection algorithms. By generating anchor boxes of varying sizes around the hand and detecting objects within these boxes, we extract objects that are in close proximity to the human's hand at points of interaction. While there may be frames where the human hand is in proximity to multiple objects, we hypothesize that the majority of the frames will solely contain the human's intended object. Therefore, by finding the object that most commonly appears in contact frames, we can accurately identify the intended object ($tag$).

To extract the object's trajectory, we use the hand interactions detected by the 100DOH model to identify frames where the human's hand is in contact ($c_t$) with an object. We then create anchor boxes of varying sizes, similar to region proposal networks in image detection \cite{ren2015faster}. Using these anchor boxes we extract objects that are in close proximity to the human's hand at points of interaction. While there may be frames where the human hand is in proximity to multiple objects, we hypothesize that the majority of the frames will only contain the human's intended object ($tag$).

% Once the object of interest is identified, we proceed to generate its trajectory ($\zeta^o_h$) in the video demonstration. Utilizing our object detector, we generate bounding boxes around the object's location for all frames in the video. With the object's location obtained in pixel coordinates ($p^o_{x_t}, p^o_{y_t}$), we apply the same de-projection technique used in hand trajectory extraction and use the depth information ($\delta_t$) to translate two-dimensional image frame coordinates into three-dimensional world coordinates ($x^o_t, y^o_t, z^o_t$). This process not only identifies the manipulated object but also delineates its trajectory throughout the video. Refer to \fig{prior} and \alg{obj_traj} for a summary.

Once the object is identified, we generate its trajectory ($\zeta^o_h$) in the video demonstration. We accomplish this by generating bounding boxes around the intended object in pixel coordinates ($p^o_{x_t}, p^o_{y_t}$). We then apply the same de-projection technique used in hand trajectory extraction and use the depth information ($\delta_t$) to translate two-dimensional image frame coordinates into three-dimensional world coordinates ($x^o_t, y^o_t, z^o_t$). This process outputs a $3$D trajectory of the intended object, capturing its movement throughout the demonstration.  Refer to \fig{prior} for a summary.

Overall our extracted prior provides us with three key pieces of information: a condensed trajectory representing the human's visited waypoints ($\xi^h$), a label identifying the object of interest ($tag$), and a trajectory indicating the object's movement ($\xi^o_h$), as summarized in Algorithm \ref{alg:obj_traj} in Appendix \ref{app:algorithms}.

\subsection{Agent-Agnostic Rewards} \label{sec:view_rewards}

After we get the human's hand trajectory ($\xi^h$) from the video demonstration, the robot executes this trajectory in the environment to try and solve the task (i.e., the human's hand trajectory becomes the robot's initial trajectory). However, this trajectory almost always fails because of morphological differences and sensor noise. In order to improve the initial trajectory over repeated interactions, the robot explores around $\xi^h$ to find the correct waypoints that solve the task (see \fig{method}). We will describe this exploration in detail in Section~\ref{sec:view_exploration}. But before we get to the exploration, we first need a feedback mechanism that allows the robot to differentiate between ``good" and ``bad" waypoints. More specifically, we design a reward model that compares how the robot is  manipulating the target object to how the human manipulated the same object during their video demonstration.

Our prior from Section~\ref{sec:view_prior} contains the $tag$ identifying the target object and its trajectory throughout the demonstration video. Similarly, we can take videos of the robot's interactions in the environment and extract the actual trajectory of the target object using the same procedure. To compare the movement of the object for the two agents, we take the mean square error (MSE) between the corresponding waypoints in their respective trajectories. The negative of this distance serves as our reward. For clarity, let the object trajectory from the prior $\xi^o_h$ consist of waypoints $(p^h_{x_{t}}, p^h_{y_{t}})$ and the object trajectory from the robot interaction $\xi^o_r$ consist of waypoints $(p^r_{x_{t}}, p^r_{y_{t}})$. Then, the reward corresponding to each waypoint is given as: 

\begin{align}
    r_{t} &= -\mid\mid \omega^r_{t} - \omega^h_{t} \mid\mid  \label{eq:reward} \\
    \omega_{t} &= (p_{x_{t}}, p_{y_{t}})
\end{align}

% We note that we measure the distance in pixels to mitigate any inaccuracies caused by transforming from the camera coordinate frame to the robot coordinate frame. Intuitively, our reward procedure is \textit{agent-agnostic} because it does not matter who is manipulating the objects --- either human or robot. We extract the object trajectories across videos from both agents, and then contrast those trajectories to quantify how similar the robot's behavior is to the human's behavior.

To minimize transformation errors, we measure this distance in pixels. This approach is agent-agnostic: we extract object trajectories from both agents' videos and compare them directly, allowing the robot to align its behavior closely with that demonstrated by the human.

\subsection{Exploration for Iterative Improvement}\label{sec:view_exploration}

% Now that we have a metric for quantifying the robot's performance, we are ready to iteratively improve the robot's trajectory. Referring back to \fig{method}, the robot starts by executing its initial trajectory extracted from the human's hand movement, and then gradually improves this trajectory by exploring around the trajectory waypoints. A typical task entails grasping an object and then manipulating that object in the same way as the human. However, the robot cannot learn about this manipulation until it has learned how to grasp the object: the successful completion of the task is contingent on the robot moving the correct object. In our running example of teaching the robot to pick up a cup, the robot cannot succeed if it grabs the wrong object (e.g., picks up a plate), or if the robot does not grasp the object securely (e.g., drops the cup). We therefore  divide the overall exploration into two parts: \textit{grasp}, where the robot finds the grasp location for the object, and \textit{task}, where the robot learns to imitate how the human manipulates that object.

With a metric to assess the robot’s performance, we can iteratively refine its trajectory. Starting with the initial trajectory derived from human hand movements (\fig{method}), the robot gradually improves this trajectory by exploring around the waypoints. In typical tasks, the robot first grasps an object and then manipulates it. Therefore, the task success depends on first establishing a secure grasp. In our running example of teaching the robot to pick up a cup, the robot cannot succeed if it grabs the wrong object (e.g., picks up a plate), or if the robot does not grasp the object securely (e.g., drops the cup). We therefore  divide the overall exploration into two parts: \textit{grasp}, where the robot finds the grasp location for the object, and \textit{task}, where the robot learns to imitate how the human manipulates that object.

Formally, the initial trajectory extracted from the human's video consists of a set of $n$ waypoints and their corresponding contact information $\xi^h = \{(\omega^h_{t}, c_t) \mid t \in [t_1, t_2, \dots, t_n] \}$ where each waypoint is a tuple $(x, y, z)$ and $c$ is the contact. Here $x, y, z$ indicate the 3D position of the hand and $c$ indicates if the hand is in contact with any objects. From this contact information we can determine when the human grasps and releases objects. For instance, let the waypoint where the contact begins be $\omega^h_{grasp}$. We use this point to divide the prior into two trajectories: $\xi^h_{grasp} = \{ (\omega^h_{t_1}, c_{t_1}), \dots, (\omega^h_{grasp + 1},$ $c_{grasp + 1}) \}$ and $\xi^h_{task} = \{ (\omega^h_{grasp + 1}, c_{grasp + 1}), \dots, (\omega^h_{t_n}, $ $c_{t_n}) \}$. 
For clarity, from now we will denote trajectories as a set of waypoints $\omega$, however, each element of the trajectory is a tuple of a waypoint and its corresponding contact $c$. 
Under VIEW, the robot separately explores around these two trajectories to pick up the object and then perform the task. Below we discuss exploration strategies for iteratively improving \textit{grasp} and \textit{task}.

% More formally, the initial trajectory extracted from the human's video consists of a set of $n$ waypoints $\xi^h = \{\omega^h_{t} \mid t \in [t_1, t_2, \dots, t_n] \}$ where each waypoint is a position and contact tuple $(x, y, z, c)$. Here $x, y, z$ indicate the 3D position of the hand and $c$ indicates if the hand is in contact with any objects. From this contact information we can determine when the human grasps and releases objects. For instance, let the waypoint where the contact begins be $\omega^h_{grasp}$. We use this point to divide the prior into two trajectories: $\xi^h_{grasp} = \{ \omega^h_{t_1}, \dots, \omega^h_{grasp + 1} \}$ and $\xi^h_{task} = \{ \omega^h_{grasp + 1}, \dots, \omega^h_{t_n} \}$. Under VIEW, the robot separately explores around these two trajectories to pick up the object and then perform the task. Below we discuss the robot's exploration strategies for iteratively improving \textit{grasp} and \textit{task}.

\subsubsection{Correcting the Grasp Waypoint}\label{sec:view_grasp}
In our first phase the robot explores around the prior $\xi^h_{grasp} = \{ \omega^h_{t_1}, \dots, \omega^h_{grasp}, \omega^h_{grasp + 1} \}$. Although this prior contains multiple waypoints, the primary focus is on $\omega^h_{grasp}$, the waypoint where it should grasp the target object. 
%How the robot arm reaches for that object is irrelevant, so long as it is able to successfully pick up and hold the item. 
The approach used to reach the object is less important as long as the grasp is successful. Accordingly, in VIEW the robot uses the position where the human grasped the object as a prior (i.e., $\omega^h_{grasp}$), and then the robot intelligently explores around this prior to find a grasp location that is effective for the robot arm and gripper.

\begin{figure}
    \centering
    \includegraphics[width=0.9\columnwidth]{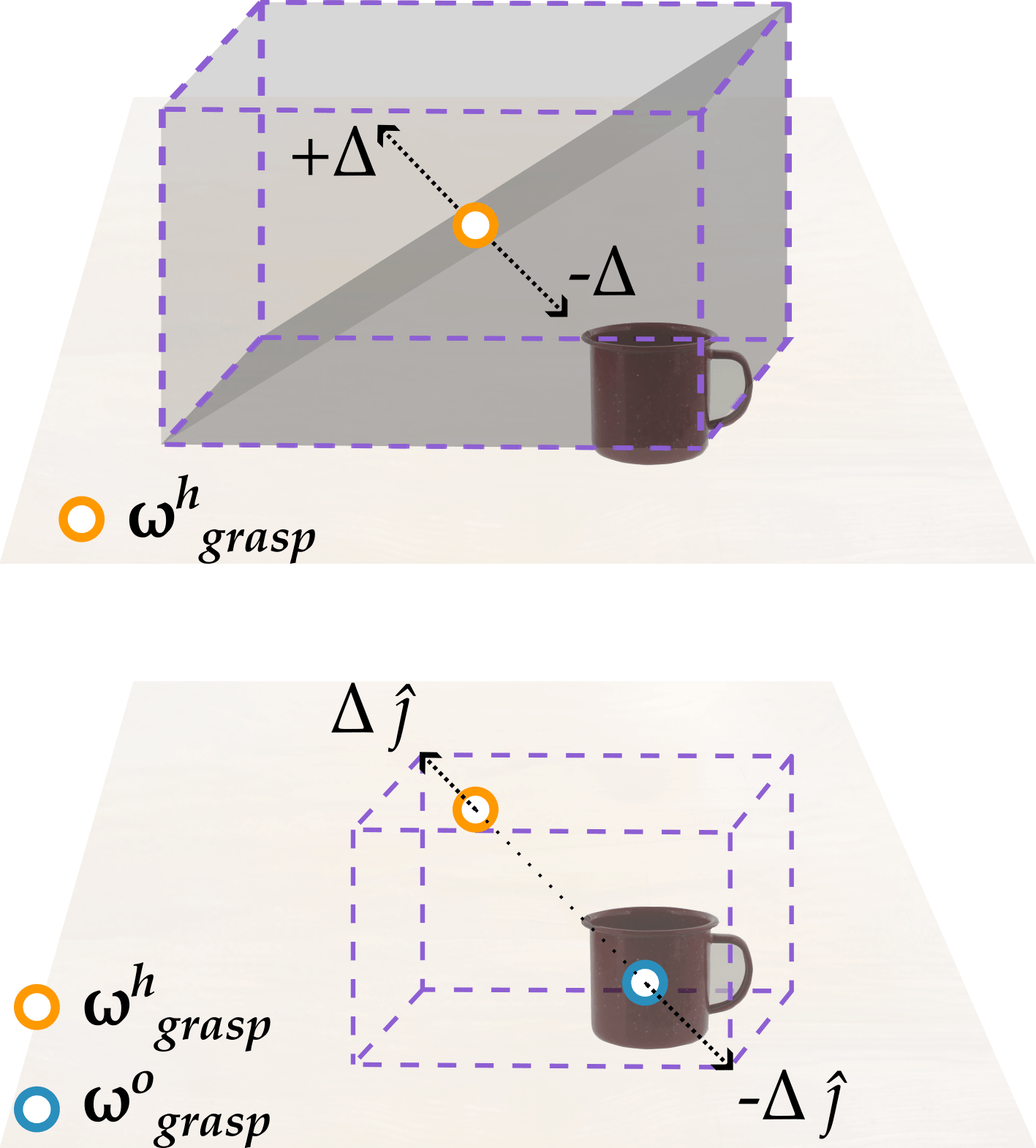}
    \caption{Generating a bounding box for exploring grasp locations. We define a region around the waypoint $\omega^h_{grasp} = (x, y, z)$ where the human first interacted with the object in the video demonstration. (Top) A naive approach: the bounding box is centered around $\omega^h_{grasp}$ with limits $\Delta$. The principal diagonal of the bounding box is defined by $(x - \Delta, y - \Delta, z - \Delta)$ and $(x + \Delta, y + \Delta, z + \Delta)$. (Bottom) Our approach that leverages the estimated object location $\omega^o_{grasp}$ at the time of grasping to bias the search space. The principal diagonal of the bounding box are $\omega^h_{grasp} + \Delta \hat{j}$ and $\omega^o_{grasp} - \Delta \hat{j}$, here $\hat{j}$ is the unit vector parallel to the principal diagonal. This bounding box is typically smaller and is more likely to include an effective grasp location for the robot. }
    \label{fig:exploration_bb}
\end{figure}

\p{Restricting the Exploration Space} We define a search region around the waypoint $\omega^h_{grasp}$ using a bounding box $\mathcal{B}$. A simplistic approach centered on $\omega^h_{grasp}$ would include an area too large for efficient search. More explicitly, defining a range in the robot's coordinates with limits $\Delta$: from $(x - \Delta, y - \Delta, z - \Delta)$ to $(x + \Delta, y + \Delta, z + \Delta)$, would create a bounding box with the waypoint $\omega^h_{grasp}$ at the center. However, such a bounding box may be unnecessarily large and span irrelevant parts of the robot workspace. In our running example of learning to pick up a cup, this bounding box could include part of the workspace which is farther away from the cup, as shown \fig{exploration_bb} (Top). Instead, we create a compact bounding box around both $\omega^h_{grasp}$ and the object location $\omega^o_{grasp}$, extending the diagonal between these points by a limit $\Delta$ to account for sensor or model inaccuracies. This bounding box, defined from $(\omega^o_{grasp} - \Delta \hat{j})$ to $(\omega^h_{grasp} + \Delta \hat{j})$, uses $\hat{j}$ as the unit vector between the two points. This focused region, shown in \fig{exploration_bb} (Bottom), allows for a more efficient search by targeting relevant areas.
% To see an example of this bounding box refer to \fig{exploration_bb} (Bottom). Intuitively, this bounding box is a more efficient search area because it is based on both the human's hand position and the estimated object position.

% \p{Rewards for Grasp Exploration} 
% Now that the robot has an exploration region $\mathcal{B}$, the robot can move to different waypoints inside that region to try and grasp the object. However, when performing this exploration a key question emerges: how can we determine if the robot's grasp was successful? Merely observing the object's location at the moment of grasping is not sufficient, as the object does not move from its initial location until it has been both grasped \textit{and} moved.

% Therefore, to assess whether the robot has grasped the desired item, we incorporate the subsequent waypoint ($\omega^h_{grasp + 1}$) into our grasp exploration trajectory. This allows the robot to execute a grasp at the chosen location and then proceed to the next waypoint along its initial trajectory. Throughout each round of exploration in the environment the robot stores its end-effector location, and we use the video camera to track the \textit{tagged} object using our detection model. Intuitively, we can confirm that the robot has successfully grasped the \textit{tagged} object if the object is positioned close to the robot's end-effector at timestep $grasp + 1$. This ensures that the grasp's effectiveness is measured not just by proximity, but also by whether the robot can hold and move the object.

\p{Rewards for Grasp Exploration}
Once the exploration region $\mathcal{B}$ is established, the robot tests different waypoints within it to find an effective grasp. A grasp is only successful if it enables the robot to pick up and move the object. To verify this, we include the next waypoint $\omega^h_{grasp + 1}$ in the exploration sequence, allowing the robot to grasp at a chosen location and proceed to the following waypoint. Since we know the object of interest ($tag$), we can confirm that a grasp is successful if the object remains close to the robot’s end-effector at timestep $grasp + 1$.

\begin{figure*}
    \centering
    \includegraphics[width=2\columnwidth]{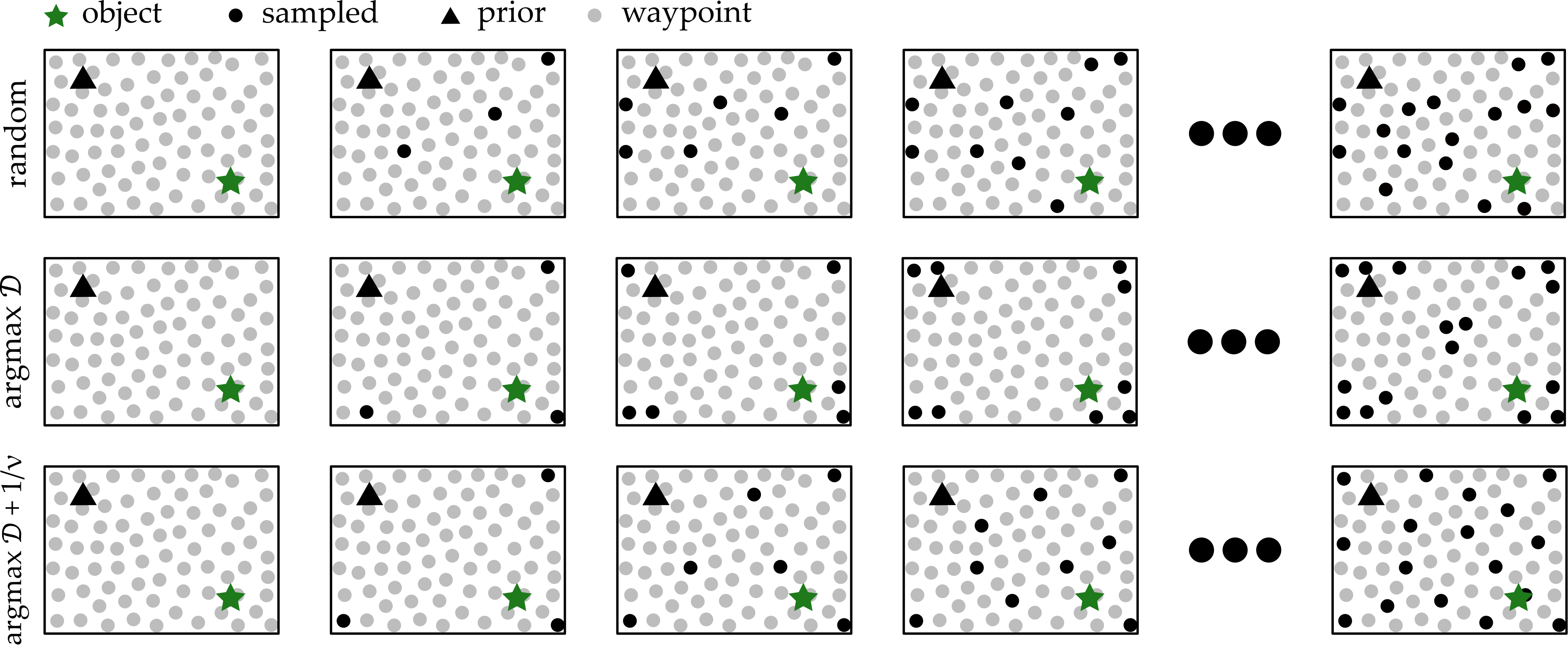}
    \caption{Comparison of different sampling methods in our \textit{high-level grasp exploration}. We show an example task in a two-dimensional space which is bounded around the prior (black triangle) and the object (green star). (Top) Each new high-level waypoint point is uniformly randomly sampled from our set of unvisited waypoints. This method can eventually reach the object with sufficient exploration. However, new samples my be close to previously tested points. (Middle) To quickly reduce the uncertainty about the unknown object location, we can sample high-level waypoints that maximize the distance to all previously visited waypoints. We expect that these waypoints will explore new regions of the search space. In practice, however, the distance-based estimation from \eq{distance_metric} results in points that are clustered at the corners and center. (Bottom) Our proposed solution is to add a regularizing term in \eq{sampling_strategy} to ensure that the next high-level waypoint is truly from an unexplored region of workspace. Our experiments show that this approach finds the grasp location more rapidly.}
    \label{fig:exploration_sampling}
\end{figure*}

% We have established where the robot should search and how to determine if a proposed waypoint has grasped the item. Our final step is developing a method for \textit{exploring} the bounding box to identify an optimal grasp location. This search problem is complicated by two main challenges. First, for most waypoints the robot does not move the object and the rewards are constant. Put another way, we have \textit{sparse rewards}. Second, if the robot reaches a waypoint that is close to the object it may accidentally knock the object over or otherwise \textit{lower its measured rewards}. Hence, waypoints that are actually close to a successful grasp could be penalized by the reward model.

\p{Grasp Exploration} The final step involves searching for the optimal grasp location within $\mathcal{B}$. This search is complicated by two main challenges. First, for most waypoints the robot does not move the object and the rewards are constant. Put another way, we have \textit{sparse rewards}. Second, if the robot reaches a waypoint that is close to the object it may accidentally knock the object over or otherwise \textit{lower its measured rewards}. Hence, waypoints that are actually close to a successful grasp could be penalized by the reward model.

Returning to our cup example, consider a scenario where the robot receives a baseline reward of $+10$ at waypoints that don’t involve moving the cup. If the cup is supposed to be moved left as per the human's video demonstration, and the robot picks a point that hits the cup and moves it to the right, the reward might drop to $+5$. Conversely, moving the cup correctly to the left might increase the reward to $+15$. In both cases, the robot has gained valuable information: the chosen waypoint interacted with the object and may be near to a successful grasp location. This variation in rewards --- regardless of whether the reward is increasing or decreasing --- helps to pinpoint the object’s location within the search space $\mathcal{B}$.

Put together, the sparse rewards at grasp locations and locally varying rewards around those locations make it difficult for the robot to efficiently optimize for successful grasps. We therefore propose a quality-diversity (QD) approach for intelligently searching the space $\mathcal{B}$. Our proposed QD algorithm is broken into a \textit{high-level} search --- which divides $\mathcal{B}$ into regions of interest --- and a \textit{low-level} search --- which explores within those regions to pinpoint a successful grasp location. We summarize this overall method in \alg{grasp_exp}.

\smallskip

% \noindent \textit{High-Level Grasp Exploration.}
% We first discretize the bounding box $\mathcal{B}$ into a set of regions for the robot to explore. To obtain these regions we use Centroidal Voronoi Tessellation (CVT) \cite{vassiliades2017using}: we numerically sample a large number of points inside $\mathcal{B}$, and then use k-means to group these sampled points into $M$ clusters. 
% In practice, these clusters provide $M$ regions that are equally spread across the bounding box $\mathcal{B}$. Within our high-level exploration process the robot will determine which of these clusters are of interest --- i.e., which clusters could contain a successful grasp location --- for more targeted low-level optimization.

% Let the centroids of the high-level clusters form a discrete set of potential waypoints: $\Omega_{unvisited} = \{\omega \quad \forall i = 1, 2, \dots M\}$. With no additional information to differentiate them, all centroids are considered \textit{equally likely} grasp locations. A straightforward method to select waypoints would be to uniformly sample from this set of univisited centroids ($\Omega_{unvisited}$). However, random sampling from a uniform distribution may lead to waypoints that are close to previously tested centroids (see \fig{exploration_sampling} Top). We therefore propose a sampling strategy for selecting waypoints from $\Omega_{unvisited}$ that encourages the robot to visit previously unexplored parts of the bounding box $\mathcal{B}$.

\noindent \textit{High-Level Grasp Exploration.}
To efficiently explore the bounding box $\mathcal{B}$, we discretize it into distinct regions using Centroidal Voronoi Tessellation (CVT) \cite{vassiliades2017using}. By numerically sampling points within $\mathcal{B}$ and applying k-means clustering, we generate $M$ evenly distributed clusters, each representing a region for exploration. The centroids of these clusters form the set of potential high-level waypoints $\Omega_{unvisited} = \{\omega \mid i = 1, 2, \dots, M\}$. The robot’s task is to identify regions likely to contain a successful grasp location for targeted refinement.

% Under our approach the robot reasons over previously attempted centroids to select a new high-level waypoint that is different from the ones it has already tested. This can be achieved by choosing a waypoint from the unvisited centroids ($\Omega_{univisted}$) that maximizes the distance from already visited waypoints ($\Omega_{visited}$). Mathematically, we optimize the selection of the next waypoint as follows:

A naive approach would be to uniformly sample waypoints from $\Omega_{unvisited}$. However, this risks selecting points close to already tested centroids, leading to redundant exploration (see \fig{exploration_sampling} Top). Instead, we propose a sampling strategy that prioritizes unexplored regions by maximizing the distance between new waypoints and previously visited ones. We define the next high-level waypoint $\omega_{next}$ as:
\begin{align}
\omega_{next} = \argmax_{\omega \in \Omega_{unvisited}} \quad  \mathcal{D}(\omega, \Omega_{visited}) \\ \label{eq:distance_metric}
\mathcal{D}(\omega, \Omega_{visited}) = \frac{1}{k} \sum_{\omega_j \in N^k_i} \mid\mid \omega - \omega_j \mid\mid
\end{align}
Here, $\mathcal{D}(\omega, \Omega_{visited})$ is the mean distance between each unvisited waypoint and its $k$-nearest neighbors ($N^k_i$) in the visited set $\Omega_{visited}$, i.e., the set of tested centroids. This ensures that new waypoints are chosen based on their separation from already tested locations.

To address potential clustering issues where a waypoint is close to one visited point but far from others (see \fig{exploration_sampling} Middle), we introduce a regularization constraint. By calculating the variance $\nu$ of distances between an unvisited waypoint and all visited waypoints, we ensure equidistance. Our final optimization balances diversity and uniformity:
\begin{equation}
    \omega_{next} = \argmax_{\omega \in \Omega_{unvisited}} \quad  \mathcal{D}(\omega, \Omega_{visited}) + \frac{1}{\nu} \label{eq:sampling_strategy}
\end{equation}

This optimization ensures that the selected waypoint from $\Omega_{unvisited}$ maximizes distance from all waypoints in the set while attempting to be equidistant with the waypoints in $\Omega_{visited}$ (see \fig{exploration_sampling} Bottom). In practice, the robot selects a high-level waypoint from $\Omega_{unvisited}$ using \eq{sampling_strategy}, and then executes a trajectory in the environment that attempts to grasp the object at that waypoint. We use the reward model from \eq{reward} to assess the performance of this grasp.

\medskip

\noindent \textit{Low-Level Grasp Exploitation.}
To prioritize regions for refinement, the robot selects waypoints from the visited waypoint set ($\Omega_{visited}$) based on the magnitude of reward changes. Waypoints where rewards varied significantly --- either increasing or decreasing --- are more likely to be close to an optimal grasp. We sample a waypoint $\omega_{local}$ from $\Omega_{visited}$ with the probability distribution:
\begin{equation}
    p_i = \frac{e^{\gamma \sigma_i}}{\sum e^{\gamma \sigma_j}} \label{eq:p_sample_visited}
\end{equation}
where the denominator is summed across all visited waypoints, and $\sigma_i$ is the normalized variation in reward between the high-level waypoint $i$ and the reward $R_{0}$ from the initial trajectory:
\begin{equation}
    \sigma_i = \frac{\mid\mid R_i - R_{0} \mid\mid}{\max_{j} \mid\mid R_j - R_{0} \mid\mid}
\end{equation}
This approach biases the robot's search toward high-level waypoints that showed the most significant changes in reward, ensuring focus on areas likely to contain a successful grasp location.

Once a high-level waypoint $\omega_{local}$ is selected, the robot defines a smaller bounding box $\mathcal{B}_{local} \subset \mathcal{B}$ centered around $\omega_{local}$ with a radius $\epsilon$. This refined region narrows the search to the immediate vicinity of the promising waypoint. Within $\mathcal{B}_{local}$, the robot uses an optimization algorithm, such as Bayesian optimization (BO)\footnote{VIEW is not tied to a specific local optimization algorithm. While we use BO in our experiments, it can be replaced with any other optimizer.} \cite{BO}, to maximize the reward function. BO iteratively samples waypoints $\omega_{opt}$ from $\mathcal{B}_{local}$, evaluating each based on the reward model (\eq{reward}). If a sampled waypoint $\omega_{opt}$ achieves a higher reward than the current $\omega_{local}$, it is added to $\Omega_{visited}$ as the new candidate for refinement.

This two-step process --- targeting promising regions through reward-based sampling and optimizing locally using BO --- enables the robot to precisely identify and finetune the optimal grasp location within the bounding box $\mathcal{B}$. Algorithm \ref{alg:grasp_exp} in Appendix \ref{app:algorithms} summarizes this exploration scheme.

\smallskip

\noindent \textit{Trading-off Between High- and Low-Level Search.}
Our overall exploration process for identifying a successful grasp trades-off between testing new high-level waypoints from $\Omega_{unvisited}$ and then exploiting the regions around relevant waypoints from $\Omega_{visited}$. We balance this exploration of new regions and exploitation of sampled regions using probability $p_{explore}$: with probability $p_{explore}$ we test an $\Omega_{unvisited}$ waypoint, and with probability $1 - p_{explore}$ the robot explores the region around a waypoint from $\Omega_{visited}$.
% Our overall exploration process for identifying a successful grasp trades-off between testing new high-level waypoints from $\Omega_{unvisited}$ and then exploiting the regions around relevant waypoints from $\Omega_{visited}$. We balance this exploration of new regions and exploitation of sampled regions using probability $p_{explore}$. Looking at Algorithm~\ref{alg:grasp_exp}, with probability $p_{explore}$ we test an $\Omega_{unvisited}$ waypoint, and rollout a trajectory in the environment that includes that waypoint. Similarly, with probability $1 - p_{explore}$ the robot executes a trajectory that explores the region around a waypoint from $\Omega_{visited}$. 
The value of this probability is chosen based on the variance in the rewards of the high-level waypoints. Concretely, the exploration probability is calculated as $p_{explore} = \frac{\alpha}{\sigma_{\max}}$ where $\sigma_{\max}$ is the highest normalized variance in reward between the high-level waypoints and $\alpha$ is a hyperparameter. Intuitively, as the variance in the rewards increases, the probability of low-level exploitation increases proportionately.
This search process ends once the robot identifies a waypoint that successfully grasps the target item.

In summary, grasp exploration works in a hierarchical manner. First the robot conducts a broad search across the bounding box $\mathcal{B}$ by dividing it into $M$ evenly distributed high-level waypoints. The robot then conducts a more refined search in the vicinity of waypoints that are potentially close to the object --- i.e., waypoints that incur a high variation in reward. Overall, our grasp exploration approach has similarities to the QD algorithm CMA-ME \cite{fontaine2020covariance}. The primary novelty of our approach as compared to \cite{fontaine2020covariance} is the sampling technique used for selecting the high-level waypoints. While CMA-ME relies on randomness to select a point from $\Omega_{unvisited}$, we propose a regularized entropy metric for selecting points that are evenly spaced across the search space. In \fig{exploration_sampling} we show an example of why this high-level sampling approach is important, and how our proposed approach can more rapidly identify the grasp location. We also test this difference in our experiments.

\subsubsection{Correcting the Task Waypoints} \label{sec:view_task}

Once the robot has grasped the target object, it can now proceed to replicate how the human manipulated that object in the demonstration video. This process is more straightforward than identifying the correct grasp because here the rewards are \textit{dense}: any change in the way the robot moves the object will lead to a change in the object's position, and therefore a change in the measured rewards from \eq{reward}. Accordingly, we can use off-the-shelf optimization methods to iteratively improve the waypoints along the initial trajectory $\xi^h_{task} = \{ \omega^h_{grasp + 1}, \dots, \omega^h_{t_n} \}$ after the robot has learned to successfully grasp the target item.

Similar to our approach for grasp optimization, we start by drawing bounding boxes $\mathcal{B}$ around each waypoint in $\xi^h_{task}$. Here it is important to remember that the reward function from \eq{reward} is the distance between the object position in the human's video demonstration and the object position in the robot's task execution. Hence, the rewards associated with each waypoint are independent, and the robot can simultaneously explore and improve each task waypoint without affecting the results across other task waypoints.
We therefore conduct $n - grasp$ search processes in \textit{parallel}, one for each waypoint from $\omega^h_{grasp + 1}$ to the final waypoint $\omega^h_{t_n}$. Let us denote the robot's updated trajectory as $\xi^r_{task} = \{ \omega^r_{grasp + 1}, \dots, \omega^r_{t_n} \}$. To find an optimal waypoint $\omega_r$, the samples a point within the corresponding bounding box and then rolls-out a trajectory that moves through that point in the environment.
Here we use Bayesian optimization (although other methods are possible): for each $\omega^r_{t} \in \xi^r_{task}$, a separate instance of BO updates the robot's waypoint to better match the video demonstration. See \alg{task_exp} in Appendix \ref{app:algorithms} for a summary.

% \begin{algorithm}
% \caption{Task Exploration}
% \label{alg:task_exp}
%     \hspace*{\algorithmicindent} \textbf{Input:} Prior trajectory of task $\xi^h_{task}$
%     \begin{algorithmic}[1]
%         \State Define bounding box for each waypoint $\omega^r \in \xi^r_{task}$
%         \State Initialize a separate Bayesian optimizer \Call{BO}{} for each waypoint in task
%         \State

%         \Function{Ask}{}
%             \State Initialize an empty list $\xi^r_{task}$
%             \For{$\omega^h_i \in \xi_{task}^h$}
%                 \State Query \Call{BO}{} for $\omega^r_i$
%                 \State Add $\omega^r_i$ to $\xi^r_{task}$
%             \EndFor
%             \State \Return $\xi^r_{task}$
%         \EndFunction
%         \State
        
%         \Function{Tell}{$\xi^r_{task}, R$}
%             \For{$i = 1, \dots, n$}
%                 \State Update corresponding \Call{BO}{} with $\omega^r_i \in \xi^r_{task}, R_i \in R$
%             \EndFor
%         \EndFunction
%         \State

%         \While{\textit{task} is not successful}
%         \State Sample trajectory $\xi^r_{task} = \Call{Ask}{}$
%         \State Execute trajectory $\xi^r_{task}$ in environment
%         \State Get the reward $R$ for each waypoint in the trajectory using \eq{reward}
%         \State Inform the explorer $\Call{Tell}{\xi^r_{task}, R}$

%         \EndWhile
%     \end{algorithmic}
% \end{algorithm}

\begin{figure*}[t]
	\begin{center}
		\includegraphics[width=1.7\columnwidth]{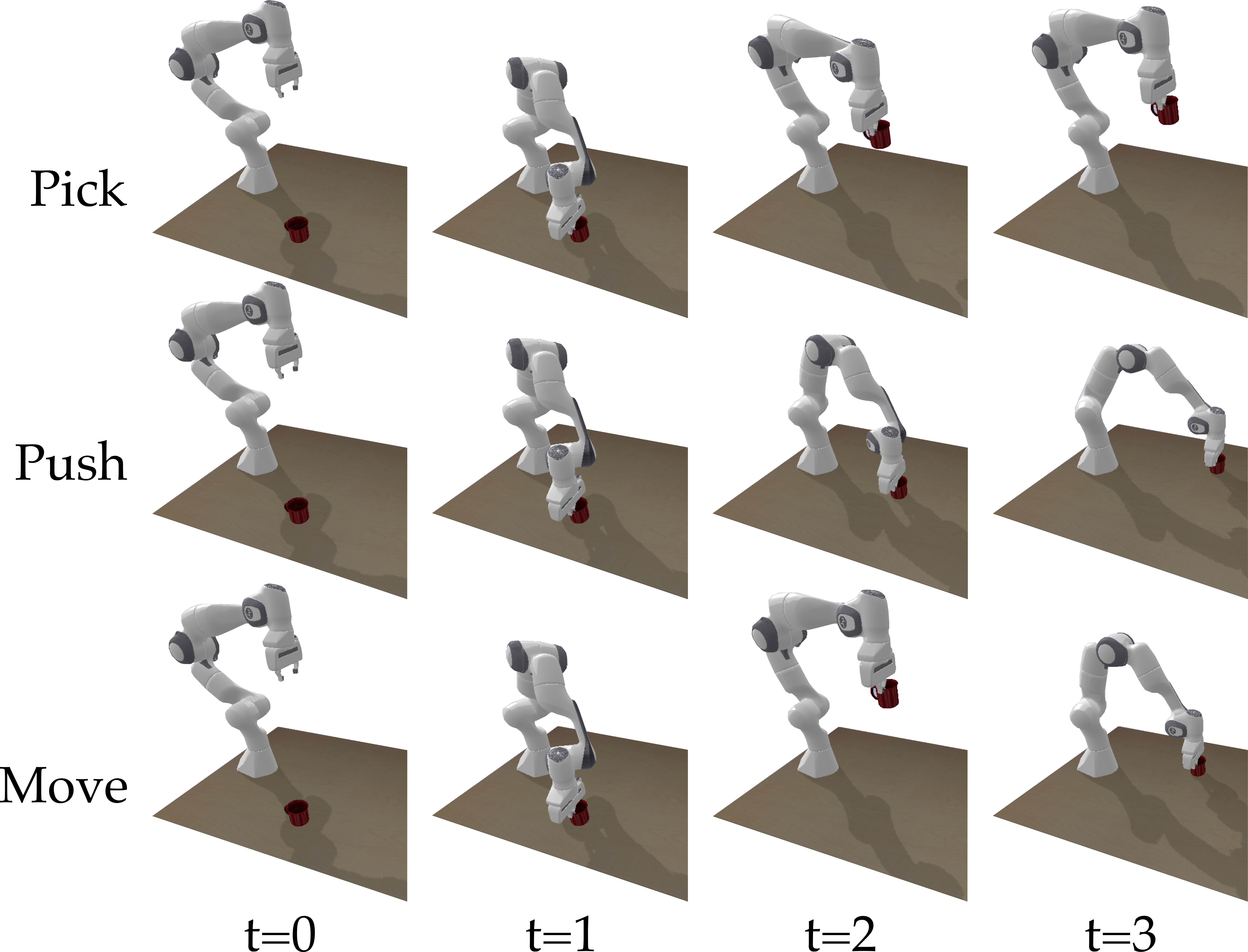}
		\vspace{-0.5em}
		\caption{Task demonstrations used in our simulations. (Top) During \textbf{Pick} the robot is teleoperated to pick up a cup placed on the table. (Middle) During \textbf{Push} the robot is teleoperated to grasp the cup and push it to a new specified location on the table. (Bottom) During \textbf{Move} the robot picks up the cup and places it at a specified new location on the table. In our ablation studies examining the effects of noise, trajectory compression, and exploration techniques, we utilize a single demonstration that is then perturbed using Gaussian noise. For assessing the influence of residual learning in our final simulation, we uniformly sample start and end points for each task and collect teleoperated demonstrations accordingly. These demonstrations are then perturbed using a deterministic noise function. We compile a dataset of $50$ demonstrations for each task, either by introducing Gaussian noise to a single trajectory or by generating $50$ distinct trajectories through uniform sampling.} 
		\label{fig:sim_tasks}
	\end{center}
	% \vspace{-2em}
\end{figure*}

\subsection{Residual Network} \label{sec:view_residual}

The process we have described so far in Section~\ref{sec:view} enables the robot to learn a task from a \textit{single} video. But when the robot gets a new video demonstration for a different task, we are faced with the question: does the robot need to restart VIEW from scratch, or can the robot leverage what it has learned on one task to accelerate learning on another task? Here we return to our example of learning to pick up a cup. Initially, the robot extracts an imperfect trajectory $\xi^h$ from the video demonstration, which it refines through exploration to arrive at a successful trajectory $\xi^*$. Ideally, the robot would have extracted $\xi^*$ directly as the initial trajectory. The discrepancy between $\xi^h$ and $\xi^*$ reflects the robot’s systematic errors during prior extraction.

We hypothesize that the error can be modeled as additive noise, expressed as $\xi^* = \xi^h + \eta_{static} + \eta_{random}$. Here, $\eta_{static}$ accounts for consistent errors --- such as sensor inaccuracies, model misalignment, or morphological differences --- that remain approximately constant across tasks. To enhance the accuracy of prior trajectory extraction, we propose training a residual network to estimate this noise. Given a dataset of $k$ previously solved tasks $\mathcal{D} = (\xi^h_k, \xi^*_k)$, we train a model to estimate the noise $\eta_{static}$. More specifically, we train a residual $\Phi(\xi^h) = \eta$ to minimize the loss $\| \xi^* - \xi^h + \Phi(\xi) \|^2$ across the dataset. The robot then deploys this residual when it receives the $k+1$ video demonstration. The robot starts by compressing the new video using the steps from Section~\ref{sec:view_prior} to get $\xi^h_{k+1}$; we then add the residual $\Phi(\xi^h_{k+1})$ to push this prior towards the correct trajectory. In practice, we will show that the residual can improve the accuracy of the prior and reduce the number of iterations the robot needs to learn new tasks.

% \textcolor{blue}{We hypothesize that some of these errors --- arising from sensor inaccuracies, model misalignment, or morphological differences --- are approximately consistent across tasks. Thus, the error can be modeled as additive noise, where $\xi^* = \xi^h + \eta_{static} + \eta_{random}$.  To improve the accuracy of prior trajectory extraction, we propose training a residual network to estimate this noise.} To de-noise the prior extraction process we propose to train a residual network across the data from previously solved tasks. Given a dataset of $k$ previously solved tasks $\mathcal{D} = (\xi^h_k, \xi^*_k)$, we train a model to estimate the noise $\eta$. More specifically, we train a residual $\Phi(\xi^h) = \eta$ to minimize the loss $\| \xi^* - \xi^h + \Phi(\xi) \|^2$ across the dataset. The robot then deploys this residual when it receives the $k+1$ video demonstration. The robot starts by compressing the new video using the steps from Section~\ref{sec:view_prior} to get $\xi^h_{k+1}$; we then add the residual $\Phi(\xi^h_{k+1})$ to push this prior towards the correct trajectory. In practice, we will show that the residual can improve the accuracy of the prior and reduce the number of iterations the robot needs to learn new tasks.

\subsection{Incorporating Multi-Object Scenarios}\label{sec:view_scaling}

Our discussions thus far have dealt with scenarios where the human interacts with a single object. However, many real-world manipulation tasks involve handling multiple objects. For example, when making tea the human might carry a cup to a specific location, and then add tea from a kettle into the cup.

The proposed method VIEW readily adapts to such multi-object scenarios. Recall that our prior extraction process outputs a hand trajectory, providing the wrist location and contact information throughout the video. We can use the changes in the contact information to divide long trajectories --- involving multiple objects --- into distinct sub-trajectories for each subtask. Each subtask then involves interaction with only one object, which we can solve using the algorithms described above.

Consider the example of making tea. By segmenting the trajectory at points where contact changes, we can create separate, manageable segments: one for moving the cup and another for adding the tea. Each subtask is then solved separately using our algorithm, which includes dividing each individual subtask into grasp and task phases and solving them using our methods in Section~\ref{sec:view_grasp} and Section~\ref{sec:view_task}. For example, we first address the cup's movement, and once complete, proceed to handle the kettle in a similar manner \footnote{See \url{https://collab.me.vt.edu/view/} for videos showcasing VIEW learning these multi-object tasks.}.Overall, this modular strategy allows the robot to systematically learn long, multi-step tasks with visual imitation learning by concentrating on one subtask at a time.

\section{Simulations} \label{sec:simulations}

\begin{figure}[t]
	\begin{center}
		\includegraphics[width=1\columnwidth]{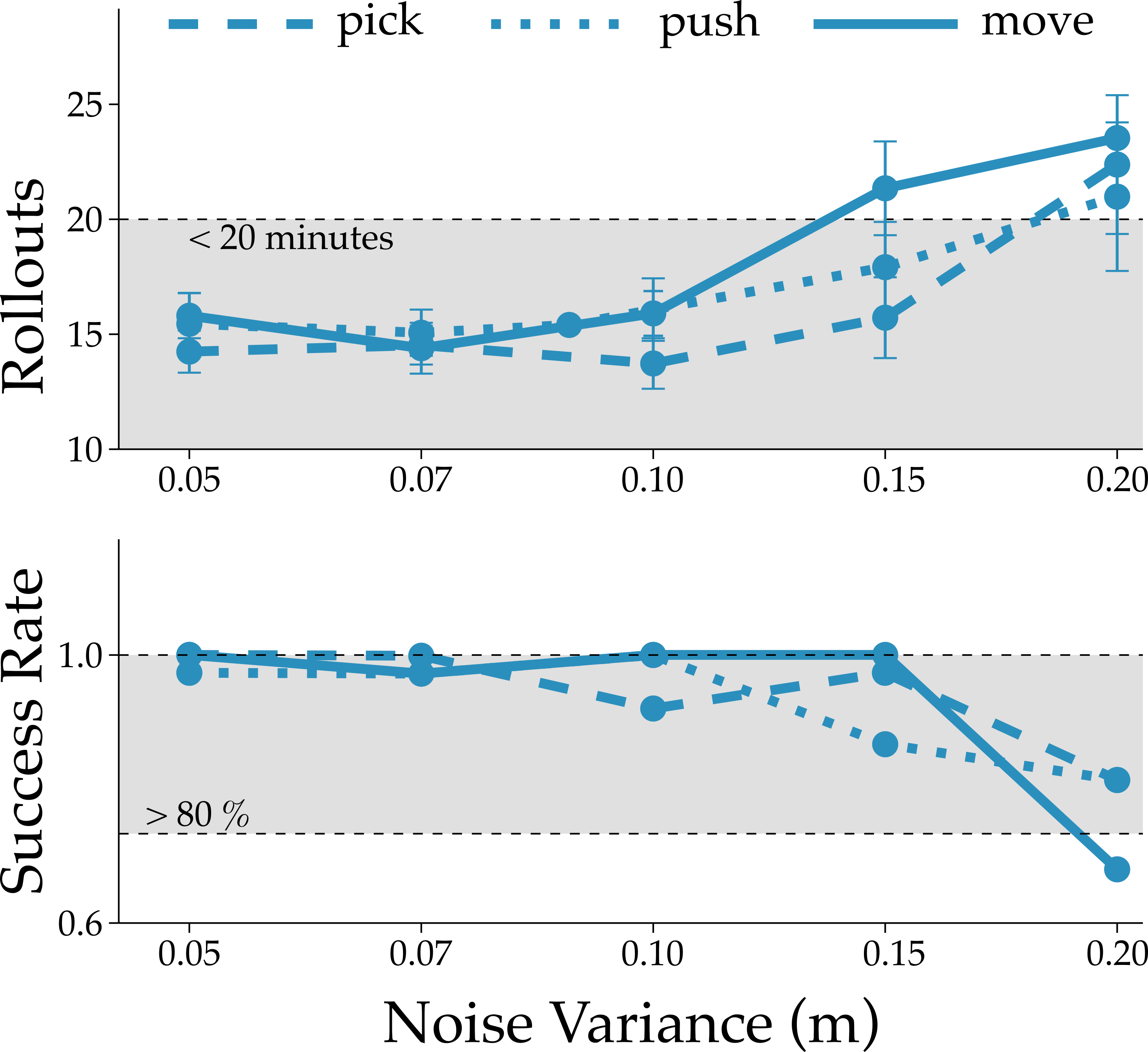}
		\vspace{-0.5em}
		\caption{Simulation results demonstrating the impact of noise on our algorithm. We test VIEW on three tasks --- pick, push, move. For each task we collect the true initial trajectory, and then add Gaussian noise to distort that trajectory. This captures scenarios where the robot's prior is incorrect (e.g., misses the cup entirely), and the robot must explore around this prior to imitate the demonstrated task. Our results are shown across $50$ trials. The shaded region in the top plot indicates less than $20$ minutes of learning time to successfully imitate the task. The shaded region in the bottom plot indicates more than $80 \%$ success rate. The bars indicate standard error of the mean.}
		\label{fig:sim_noise}
	\end{center}
	% \vspace{-2em}
\end{figure}

We proposed VIEW, a waypoint-based algorithm that can imitate humans by watching video demonstrations. We hypothesize that each component of VIEW will significantly impact the overall success of the robot. To test this hypothesis, in this section we conduct an ablation study that investigates how each part of VIEW contributes to the overall robot performance.

\p{Experimental Setup} The simulations are conducted in a Pybullet environment. To collect demonstrations, we control a simulated FrankaEmika robot arm and record frames at the rate of $20$Hz. Demonstrations are collected for three tasks: picking up an object (\textbf{pick}), pushing an object (\textbf{push}), and picking and placing an object (\textbf{move}). A single object (a cup) is used for all evaluations (See \fig{sim_tasks}). For the push and pick tasks, the initial trajectory has three waypoints after compression, while the move task yields four waypoints. To approximate noise in the real world, we distort these initial trajectories using either Gaussian noise or a fixed noise matrix. To understand our algorithm's performance in ideal conditions, no noise is injected into the reward function. The success criteria vary slightly between tasks: for \textbf{pick}, success means the robot has picked up the cup; for \textbf{push}, it is successful if it pushes the cup to the correct location; and for \textbf{move}, success requires the robot to pick up the cup and place it at the correct location on the table.

\begin{figure*}[t]
	\begin{center}
		\includegraphics[width=2\columnwidth]{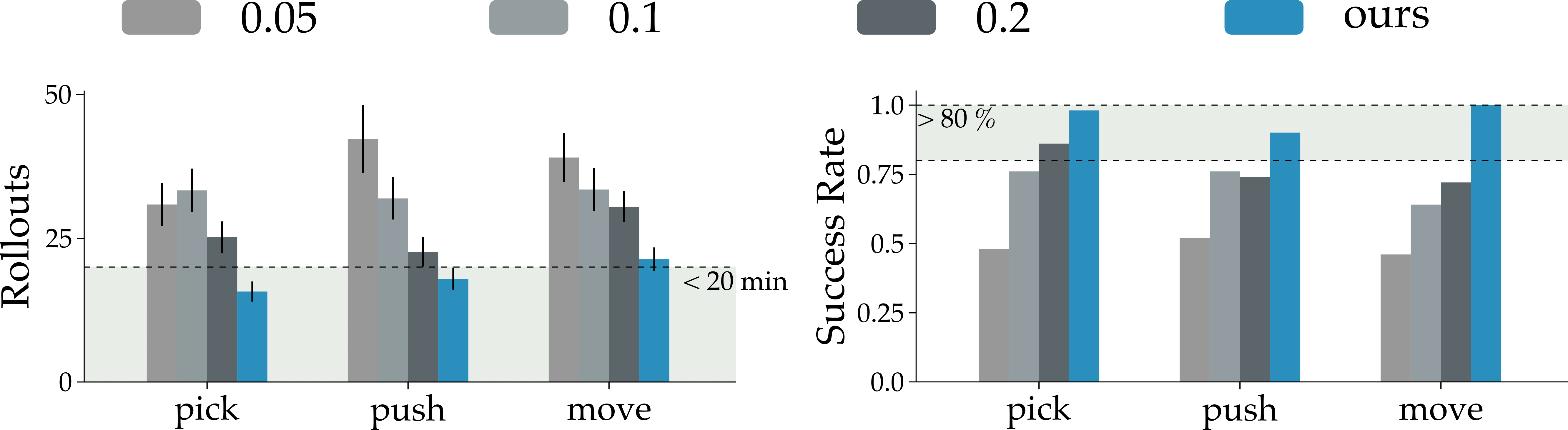}
		\caption{Simulation results examining the impact of trajectory compression. Within VIEW we compress the prior trajectory to minimize the number of waypoints while maximizing the accuracy of the compressed trajectory. We compare this method with an alternative approach in which the prior is sampled at a lower frequency to limit the number of points in the trajectory. We vary the sampling frequency of the prior trajectory to be $5$Hz, $10$Hz, or $20$Hz. The plot on the left shows the average number of rollouts it takes to learn each task over $50$ trials, and the shaded region indicates less than $20$ minutes of training time. The plot on the right shows the success rate for each task across $50$ trials, and the shaded region shows a success rate higher than $80\%$. The bars indicate standard error of the mean.} 
		\label{fig:sim_compression}
	\end{center}
	% \vspace{-2em}
\end{figure*}

\subsection{Impact of Noise}

In our first simulation, we study how noise in the extracted prior influences our algorithm's capability. Here increasing noise means that the robot's extraction of the human's hand trajectory is farther from the actual trajectory that the human followed. For each task we carry out a series of $50$ trials. Since this simulation is designed to isolate the impact of noise on our exploration scheme, we do not include the residual network during these trials.

Our findings (refer to \fig{sim_noise}) reveal that VIEW can use exploration to overcome an incorrect prior. For noise variance between $0.05$m and $0.15$m, the robot is able to successfully imitate the simulated human demonstration in almost $100\%$ of the trials. However, as the prior is distorted farther away from the correct trajectory, the performance of VIEW eventually decreases. At a noise variance of $0.2$m we observed a notable decrease in the success rate. This observation aligns with our expectation that larger distortions lead to longer search times, potentially resulting in timeouts before solutions are found. This pattern is also evident in the number of exploration rollouts required for convergence: in general, the more noise in the prior the more exploration rollouts the robot needed to correct its waypoints. Finally, we note that the number of waypoints can impact performance: tasks involving more waypoints (\textbf{move}) generally required more rollouts than tasks with fewer waypoints (\textbf{pick} and \textbf{push}). In practice, this simulation suggests that VIEW's exploration steps are critical to success, and the robot can use exploration to overcome errors in its initial guess of the correct trajectory.

\begin{figure*}[t]
	\begin{center}
		\includegraphics[width=1.4\columnwidth]{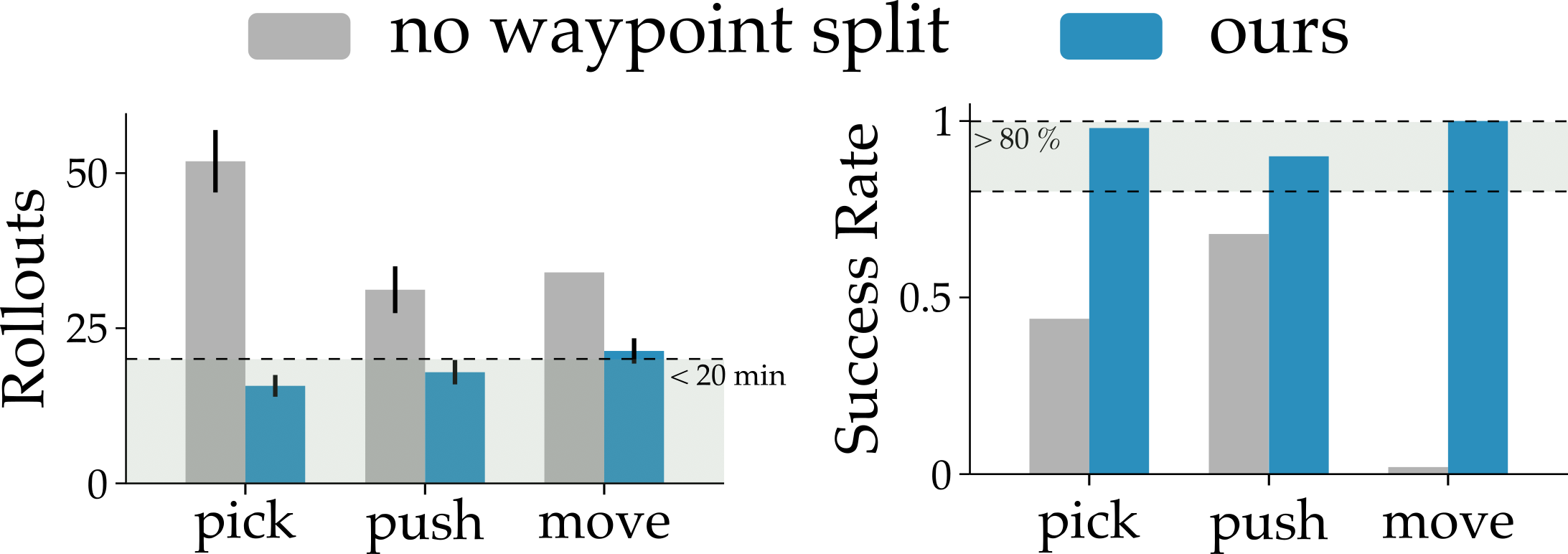}
		\caption{Simulation results examining how separating the waypoints into grasping and manipulation phases affects performance. Under VIEW the robot autonomously splits the task into separate parts: first the robot learns to grasp the object, and then it learns how to manipulate that object and complete the task. We compare this division against a unified approach that solves the entire task simultaneously. We measure the average number of rollouts taken to solve the task (Left) and the success rate (Right) over $50$ trials. The shaded regions indicate less than $20$ minutes of training time and over $80\%$ success rate, respectively. The bars indicate standard error of the mean.} 
		\label{fig:sim_exploration}
	\end{center}
	% \vspace{-2em}
\end{figure*}

\subsection{Impact of Trajectory Compression}

In our second simulation we assess the importance of trajectory compression within our algorithm. In stead of our proposed compression algorithm, simpler methods are also possible: for instance, we could simply sample the demonstration at a reduced rate, and use the sampled points as the initial trajectory (e.g., down-sample the video every $100$ frames). Here we compare the impact of this alternative method against our compression algorithm.

To generate these alternative compressions, we first distort the correct initial trajectory using a noise variance of $0.15$m. We then resample this modified demonstration at a fixed sample rate. We tested sampling rates from $20$Hz to $5$Hz. Our original demonstrations comprised approximately $40$ waypoints: hence, the compression could range from $40$ waypoints at the highest sampling rate to $10$ waypoints at the lowest sampling rate. We did not sample lower than $10$ waypoints using a fixed sampling rate because this caused the trajectory to skip critical waypoints, such as the pick point. Similar to the previous subsection, we assessed the impact of compression using the success rate and the number of rollouts required for convergence across $50$ trials.

Our results, depicted in \fig{sim_compression}, provide two important outcomes.
First, as the number of waypoints in the robot's trajectory increases (higher sampling rate), the number of rollouts required for convergence also rises. This suggests that compression is indeed important --- we can accelerate the robot's visual imitation learning by focusing on a smaller number of waypoints. Second, using simplistic compression algorithms that down-sample the demonstration at a fixed rate perform worse than our VIEW approach. 
The key difference here is that sampling at a fixed rate may cause the robot to miss a critical point along the demonstration (such as the frame where the human grasps the cup).
Using VIEW, the robot minimizes the number of waypoints, while also ensuring that those waypoints retain critical aspects of the demonstration.

\subsection{Impact of our Exploration Approach}

% In our third simulation we delve into the choice of exploration strategies. In particular, we study whether splitting the task into separate parts for grasping and manipulation is necessary. In Section~\ref{sec:view_exploration} we developed separate exploration schemes for each of these phases, with the rationale that the robot must first learn to grasp the object before it can imitate the rest of the human's demonstration. While we discussed the reasons behind this approach in Section~\ref{sec:view_exploration}, we now aim to empirically assess its effectiveness.

% As an alternative to our proposed method of task segmentation, we examine the performance of a unified optimization approach. This baseline does not separate the task into grasp and manipulation phases; instead, it performs Bayesian Optimization \cite{BO} to de-noise the \textit{entire} trajectory. We maintain the noise level at $0.15$m, and evaluate success rates and convergence rollouts for all $50$ trials, similar to our previous simulations.

In our third simulation we investigate the effectiveness of our exploration strategy. As an alternative to our proposed exploration strategy, we examine the performance of a unified optimization approach. This baselines does not separate task into grasp and manipulation phases; instead, it performs Bayesian Optimization to de-noise the \textit{entire} trajectory. We maintain the noise level at $0.15$m, and evaluate success rates and convergence rollouts for across $50$.

The outcomes of this experiment are depicted in \fig{sim_exploration}. As anticipated, we observe a substantial reduction in success rates when the waypoints are not split into grasping and manipulation phases. The highest success rate achieved without splitting is approximately $70$\% for the \textbf{push} task, whereas our segmented approach with VIEW achieves a minimum of $92$\% on the same task. VIEW also decreases the number of environmental rollouts required for convergence. Finally, we noticed that these results are impacted by the number of waypoints in the trajectory. For instance, in the \textbf{move} task --- which has four waypoints instead of the three in \textbf{pick} and \textbf{push} --- the success rate of the baseline approaches zero. These results indicate that forgoing waypoint segmentation during exploration not only leads to inferior performance, but also fails to scale effectively with an increasing number of waypoints.

\begin{figure*}[t]
	\begin{center}
		\includegraphics[width=1.4\columnwidth]{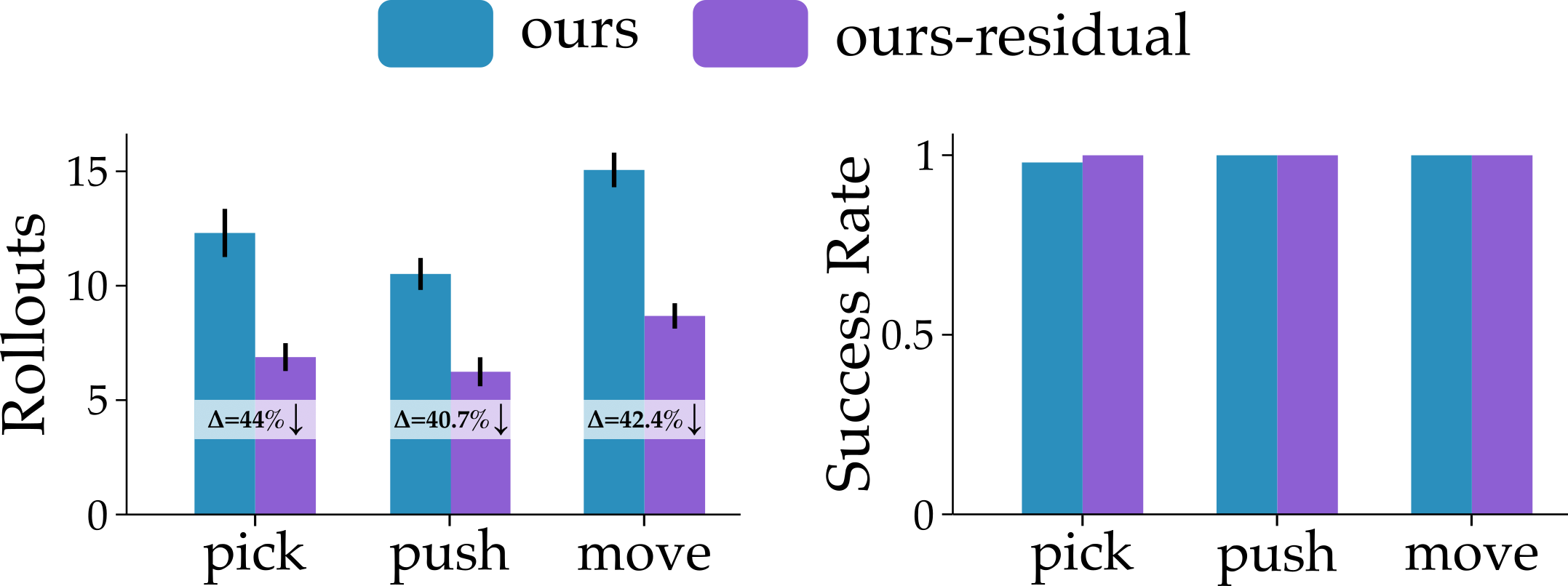}
		\caption{Simulation results for VIEW with and without the residual. We examine if the robot can utilize previous experiences to more rapidly imitate new tasks. In this simulation we sample $50$ random locations from the robot's workspace and their corresponding distortions from a noise matrix (\eq{noise_matrix}). We then use these samples to train a residual network that de-noises the distorted prior. The plots above compare the performance of our approach with and without the usingthis residual. (Left) The number of rollouts taken to solve the task averaged over $50$ trials. We also list the percentage decrease in rollouts when the residual is present. (Right) The success rate for each task. The bars indicate standard error of the mean.}
		\label{fig:sim_residual}
	\end{center}
	% \vspace{-2em}
\end{figure*}

\subsection{Impact of Residual}

% In our final simulation, we move beyond learning a single task, and explore how VIEW performs across multiple tasks. Specifically, we test how the residual network from Section~\ref{sec:view_residual} can accelerate the robot's learning on one task given that the robot has previously solved other tasks in the same environment. We contrast this to our previous simulations, where each new task or demonstration was approached from scratch.

% In the previous simulations we employed Gaussian noise and sampled various distortions for a fixed cup location. This is not feasible here because repeatedly sampling from a Gaussian distribution would cause the environmental noise to be inconsistent. Put another way, a given $xyz$ coordinate could be distorted in different ways between each task; this inconsistency would not match realistic conditions. Instead, real-world noise factors --- such as the de-projection inaccuracies and morphological differences --- impose a consistent offset at each waypoint. To better simulate these real-world conditions, we introduce a nonlinear noise matrix to distort the robot's entire workspace:

In our final simulation, we assess how the residual network can accelerate the robot’s learning, contrasting it with previous simulations where each task or demonstration was treated independently. Earlier, we used Gaussian noise to introduce various distortions, while keeping the cup’s location fixed to ensure consistent analysis --- only the robot's initial trajectory was modified. However, Gaussian noise is unsuitable here, as it would produce inconsistent offsets at each waypoint for the same object in the same location. To model a more consistent offset similar to $\eta_{stable}$ in real-world conditions, we apply a nonlinear noise matrix to introduce consistent distortions across the robot’s entire workspace:
\begin{equation}
    \eta = \tanh{\frac{\xi - \mathcal{C}}{\lambda}} \label{eq:noise_matrix}
\end{equation}
We utilize $\tanh$ to introduce distortions into the trajectory waypoints, adjusting their positions based on their proximity to a centroid ($\mathcal{C}$). The degree of distortion is modulated by the regularizer $\lambda$. This noise is then added to the demonstration to get a distorted initial trajectory $\xi^h$. We adjust the location of $\mathcal{C}$ and the value of $\lambda$ to ensure that the distortions range from $4$cm to $30$cm across all waypoints. To mitigate against any bias, we do not use a fixed cup location; instead, we sample the cup's location from a uniform distribution across the table, and then collect demonstrations for each task. We gather a total of $50$ random demonstrations from the environment, each distorted via the noise matrix, to form our dataset. Specifically, our dataset $\mathcal{D}$ for training the residual consists of $50$ pairs of initial trajectories $\xi^h$ and their corresponding ground truths $\xi^*$. Our algorithm's performance --- with and without the integration of the residual network --- is then evaluated across $50$ new and unexplored cup locations for each task.

Our results are illustrated in \fig{sim_residual}. Across all tasks, VIEW with the residual demonstrated the ability to few-shot learn new object locations. We observed a reduction of over $40$\% in the number of rollouts required for the robot to learn each task. Indeed, VIEW with the residual network needed fewer than $10$ trials on average to learn the correct behavior from distorted input trajectories. These results suggest that VIEW is not only effective when learning from scratch; we can also leverage the tasks that VIEW learned across previous video demonstrations to accelerate learning on a new video demonstration in the same workspace.

\section{Experiments} \label{sec:experiments}

% In the previous section we explored the components of VIEW through an ablation study in a simulated environment. In this section we now test our overall method in the real-world with human video demonstrations. We start by collecting video demonstrations for various tasks such as picking up a cup or moving a basket. We then apply VIEW to extract the human hand and object priors from these videos (see \fig{exp_tasks_prior}), and explore waypoints around these priors while repeatedly interacting with the environment until the robot successfully imitates the task. To see videos of these demonstrations and VIEW's learning process, visit: \url{https://collab.me.vt.edu/view/}

In the previous section we explored the components of VIEW through an ablation study in a simulated environment. In this section we now test our overall method in the real-world with human video demonstrations. We apply VIEW on video demonstrations for various tasks such as picking up a cup or moving a basket. To see videos of these demonstrations and VIEW's learning process, visit: \url{https://collab.me.vt.edu/view/}

\begin{figure*}[t]
    \centering
    \includegraphics[width=2\columnwidth]{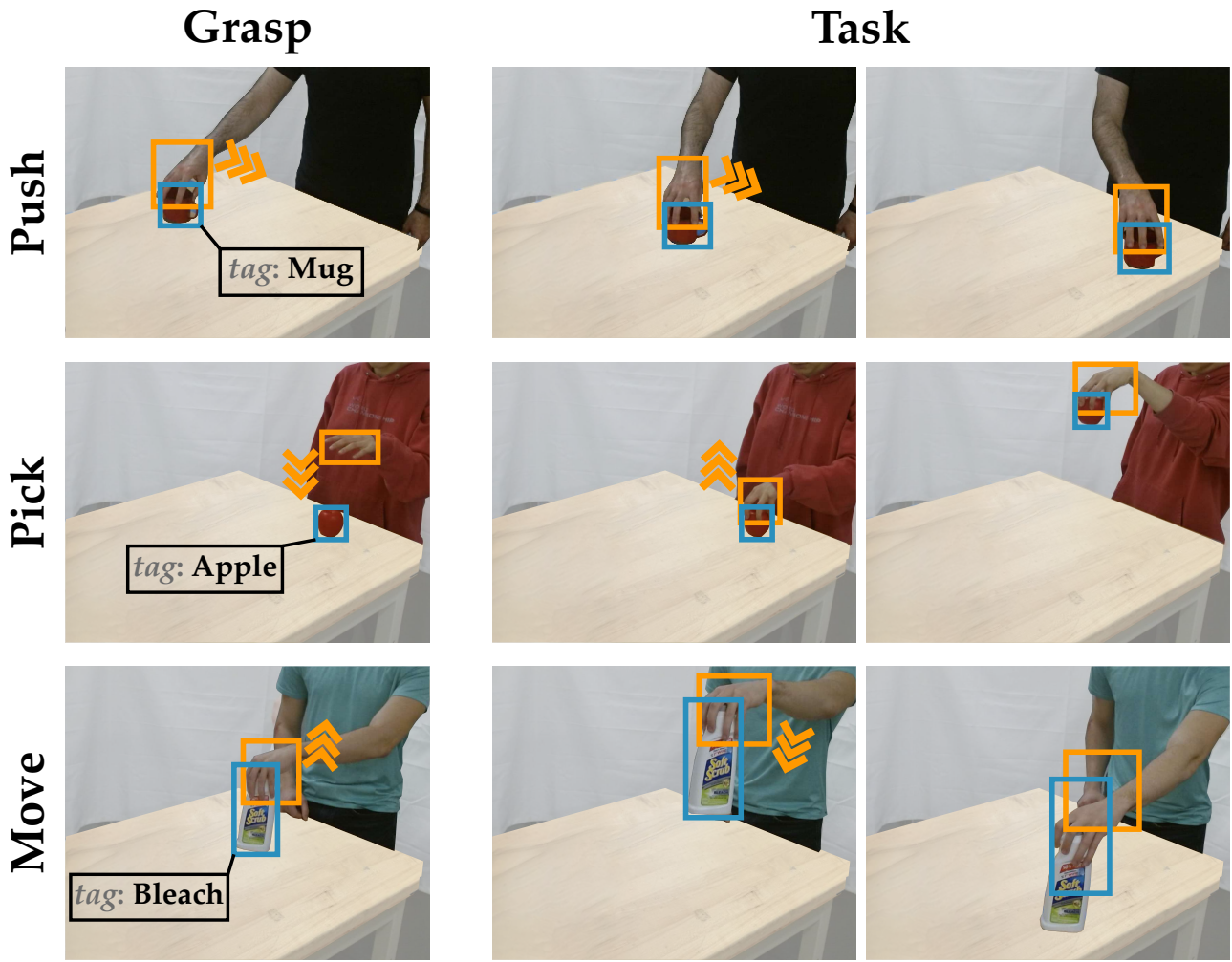}
    \caption{Manipulation tasks from our experiments. People (including the authors and external participants) provided video demonstrations of three fundamental skills necessary for more complex tasks \cite{padalkar2023open}: \textbf{push}, \textbf{pick}, \textbf{move}. Here we show examples frames where VIEW detected the human hand and the intended object, i.e., the object human is interacting with. VIEW used these frames to extract a prior trajectory for the human hand and object.}
    \label{fig:exp_tasks}
\end{figure*}

\p{Tasks} Our real-world tasks span different skills and objects. We focus on three primitive skills --- push, pick, move (\fig{exp_tasks} shows a demonstration for each skill). A full list of the objects used in our experiments is found in the Appendix: these objects include household items such as foods, cups, and containers. We start with simple tasks where the robot must learn the primitive skills in \textit{Uncluttered} environments where no other items are present. Next, we provide video demonstrations in \textit{Cluttered} settings with multiple objects, and the robot must learn to imitate the demonstrated task despite this environmental clutter.

% In \textit{Uncluttered} tasks we test the three fundamental skills. \textbf{Push}: the robot must reach for the object and push it to a randomly assigned goal position; \textbf{Pick}: the robot must learn how to pick up an object; and \textbf{Move}: the robot must reach for an object, pick it up, and place it at a randomly assigned goal location. \fig{exp_tasks} shows a demonstration for each skill.

% In \textit{Cluttered} tasks we test only the \textbf{Move} skill: here the robot must reach for and move the correct object while avoiding and ignoring the environmental clutter. Different objects may be placed close together in the environment to visually saturate the robot's camera or constrain the grasp locations for the target item.

In \textit{Uncluttered} tasks we test all three fundamental skills. However, in \textit{Cluttered} tasks we test only the \textbf{Move} skill: here the robot must reach for and move the correct object while avoiding and ignoring the environmental clutter. Different objects may be placed close together in the environment to visually saturate the robot's camera or constrain the grasp locations.

Our method can scale to arbitrarily long tasks that involve manipulating multiple objects, as shown in our supplemental videos. However, for the purposes of this experiment, we only focus on single object manipulation tasks. Our aim is to test VIEW's ability to imitate manipulation tasks from a single video demonstration, and to compare VIEW to relevant baselines.

% \p{Baselines}
% The primary baseline in our experiments is \textbf{WHIRL} \cite{whirl}, a state-of-the-art method for visual imitation learning from human demonstrations. However, the version of WHIRL implemented in our experiments differs in one way from the method described by \cite{whirl}. Within the original work, WHIRL calculates rewards by comparing agent-agnostic representations of the human demonstration and robot interaction (similar to VIEW). WHIRL finds this agent-agnostic representation by inpainting the human and the robot from the videos using Copy-Paste Networks \cite{lee2019copy}, and then using the action-recognition model of \cite{monfort2021multi} to calculate its representation. But in our experiments we found that the Copy-Paste networks could not successfully inpaint the robot, despite careful fine-tuning on a custom dataset \footnote{See the Appendix for more detailed analysis}. Accordingly, to create a fair comparison, we replaced the original reward model in WHIRL with our object-centric reward model from Section~\ref{sec:view_rewards}. We believe this is a reasonable change because our reward model actually provides more explicit feedback: it directly compares the movement of the target object across videos, rather than comparing a high-dimensional action representation as done in \cite{whirl}. The rest of the WHIRL algorithm matches the original manuscript.

\p{Baselines}
Our primary baseline is \textbf{WHIRL} \cite{whirl}, a state-of-the-art method for visual imitation learning from human demonstrations. However, our implementation of WHIRL differs slightly from the original method. Within the original work, WHIRL calculates rewards by comparing agent-agnostic representations of the human demonstration and robot interaction (similar to VIEW). WHIRL finds this agent-agnostic representation by inpainting the human and the robot from the videos using Copy-Paste Networks \cite{lee2019copy}, and then using the action-recognition model of \cite{monfort2021multi} to calculate its representation. In our experiments, however, Copy-Paste Networks could not reliably inpaint the robot, despite extensive fine-tuning on a custom dataset\footnote{See the Appendix for more details}. To ensure a fair comparison, we replaced WHIRL's original reward model with our object-centric reward model from Section~\ref{sec:view_rewards}. This substitution is reasonable as our reward model provides more explicit feedback by directly comparing target object movements rather than using high-dimensional action representations, as in the original WHIRL. The rest of the WHIRL algorithm matches the original manuscript.

Our other experimental baselines are ablations of our approach. At one extreme we have \textbf{Prior}, a method that extracts the human hand trajectory from the video demonstration and then replays that trajectory on the robot arm. This corresponds to VIEW without any exploration or residual. In practice, \textbf{Prior} will only succeed if the initial trajectory the robot extracts is sufficient to successfully imitate the task. At the other extreme we tested \textbf{ours-BO}, an ablation of VIEW that leverages a different exploration scheme. \textbf{ours-BO} is a variant of VIEW that does not use the QD algorithm: instead, the robot employs Bayesian Optimization to separately identify the grasp and match the human's behavior. Finally, when the robot is learning multiple tasks, we also test \textbf{ours-residual}. This is our full VIEW algorithm that leverages previously solved tasks to improve its prior extraction.

\p{Experimental Setup and Procedure} Experiments were conducted on a Universal Robots UR10 manipulator with $6$ Degrees-of-Freedom. The human's video demonstrations were recorded with a RealSense D435 RGB-D camera at $60$ frames per second. We also recorded the robot's interactions with the environment as it iteratively tried to imitate the demonstrated behavior.

% \begin{figure}
%     \centering
%     \includegraphics[width=1\columnwidth]{Figures/exp_tasks_prior_compressed.pdf}
%     \caption{Examples of the prior extracted from the human video demonstrations. Our method outputs how the human hand moves ($\xi^h$) and how the object moves ($\xi^o_h$) throughout the video. Each trajectory is compressed such that it only consists of waypoints that mark significant changes in the motion (like change in direction, change in contact, etc.).}
%     \label{fig:exp_tasks_prior}
% \end{figure}

We collected $13$ video demonstrations across all tasks, where $9$ were in \textit{Uncluttered} environments and $4$ were in \textit{Cluttered} environments. For the \textit{Uncluttered} tasks the $9$ total demonstrations were divided into $3$ videos for each skill --- \textit{move-uncluttered}, \textit{pick-uncluttered}, and \textit{push-uncluttered}. The $4$ videos in four different \textit{Cluttered} environments all demonstrated the same skill \textit{move-cluttered}. We conducted three trials on the robot for every demonstration, totalling $39$ trials per method.

% Between each trial \textbf{Prior}, \textbf{WHIRL}, and \textbf{ours} all reset and then learned the new task entirely from scratch. Put another way, even if the robot had learned to pick up a cup in the previous trial, the robot discarded that successful trajectory when starting the next trial. Here \textbf{ours-residual} was the exception: after we trained \textbf{ours} on the \textit{Uncluttered} tasks, we used the data from these solved tasks to train our residual policy. We then tested \textbf{ours-residual} on the four \textit{Cluttered} tasks and compared its performance to \textbf{ours} (i.e., our VIEW algorithm without including the residual).

\subsection{Results}

\begin{figure*}
    \centering
    \includegraphics[width=1.8\columnwidth]{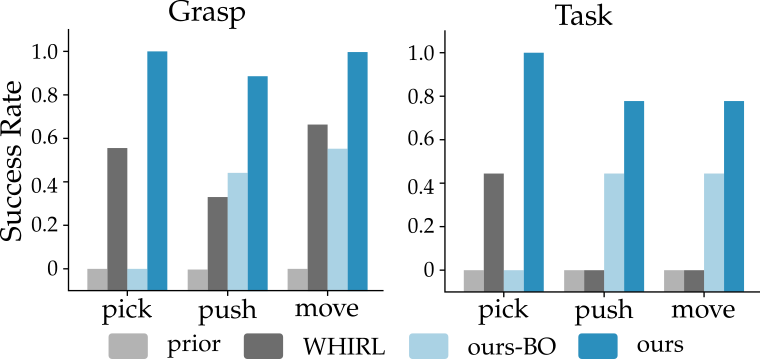}
    \caption{Experiment results for \textit{Uncluttered} tasks. (Left) How frequently the robot grasped the object from the human's video demonstration. (Right) How frequently the robot learned to imitate the human's video demonstration. Results are calculated across $9$ separate video demonstrations and $3$ trials per video demonstration. Note that the results for \textbf{pick} are the same in both grasping and task exploration, since here the objective is just to pick up (i.e., grasp) an item.}
    \label{fig:exp_single}
\end{figure*}

\p{Uncluttered Tasks} The results from our experiments on \textit{Uncluttered} tasks are shown in \fig{exp_single}. These plots display the results across the two phases of each task: the success rate for learning to grasp the object, and the success rate for learning to correctly manipulate that object. The robot is said to have succeeded in grasping if it was able to pick up the object. The definition of successful task completion varied between the different tasks: for \textit{move-uncluttered}, the task was considered a success if the robot placed the object close to the same location as the human. In \textit{Pick-uncluttered}, the robot was successful it if grasped the target object and lifted it off the table, while in \textit{push-uncluttered}, the robot successfully completed the task if it pushed the item to the human's demonstrated location.

We observed that the prior trajectory was typically distorted by 10-15 cm from the correct pick location, and simply replaying the trajectory extracted from the human's video demonstration was consistently unsuccessful. Across all tasks, \textbf{Prior} was not able to either grasp or manipulate the target object. The state-of-the-art visual imitation learning baseline \textbf{WHIRL} was more effective, particularly in learning to grasp the target item. But \textbf{our} proposed VIEW algorithm surpassed this baseline, reaching more than twice the success rate of \textbf{WHIRL} for the push task and achieving a $100\%$ success percentage in the pick task. For \textit{push-uncluttered} and \textit{move-uncluttered}, \textbf{WHIRL} was able to grasp the target object in some trials, but it did not learn to correctly manipulate that object within the limit of $100$ rollouts in the environment (roughly $45$ minutes). For these same tasks \textbf{our} VIEW algorithm reached an $80 \%$ success rate, learning to replicate the human's video demonstrations in less than $30$ minutes\footnote{WHIRL was shown to work for similar tasks in the original paper \cite{whirl}. However, we were unable to reproduce these results. We acknowledge that we replaced WHIRL's original reward model with our own agent-agnostic reward. However, this new reward provides more explicit feedback about the task and exploration. See Appendix for more details.}.

Finally, we compared the performance of \textbf{ours-BO} and \textbf{ours}. Across the board, we found that \textbf{ours-BO} is less effective at visual imitation learning than our full VIEW algorithm, and in the \textit{pick-uncluttered} environment this baseline performs significantly worse than \textbf{WHIRL}.
These results highlight the importance of our high-level and low-level QD search algorithms for exploring how to grasp the object: without the ability to learn effective grasps, \textbf{ours-BO} struggles to imitate the rest of the manipulation task.

\begin{figure}
    \centering
    \includegraphics[width=0.98\columnwidth]{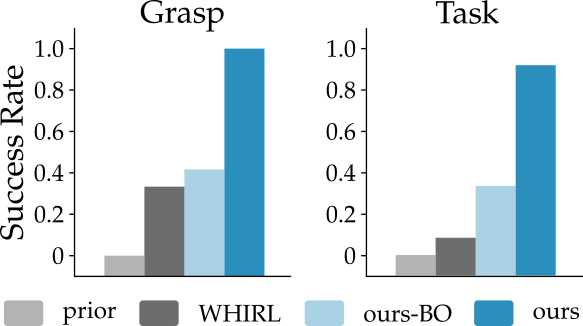}
    \caption{Experiment results for \textit{Cluttered} tasks. Here the environment contained multiple extraneous items in addition to the target object the human manipulated. (Left) How frequently the robot learned to grasp the correct item. (Right) How frequently the robot correctly imitated the entire video demonstration. These results were taken across $4$ separate video demonstrations and $3$ trials per video demonstration (for a total of $12$ datapoints).}
    \label{fig:exp_multi}
\end{figure}

\p{Cluttered Tasks}
We present the results for the \textit{Cluttered} task trials in \fig{exp_multi}. As before, the plot on the left shows the grasp success rate, and the plot on the right shows the full task success rate. Our results followed the same trends as in \textit{Uncluttered} tasks. Directly executing the extracted prior never led to success for either grasping or manipulation. The baselines \textbf{WHIRL} and \textbf{ours-BO} were roughly similar, reaching success percentages of less than $50\%$ across a maximum of $100$ real-world rollouts (roughly $45$ minutes). We were not surprised that \textbf{WHIRL} struggled with cluttered environments: it does not split the exploration into separate parts for grasping and manipulation; even if the robot grasps the object in an interaction, it can fail to grasp it again in the subsequent repetitions\footnote{It is important to note that our reward model provides explicit feedback about the \textit{tagged} object: the rewards do not change if WHIRL moves any object other than the \textit{tagged} object. In contrast, the original reward model in WHIRL compares the agent-agnostic action embeddings. This would pose a significant challenge in a cluttered environment since the robot would still be executing the right behavior, but mainpulating the wrong object. For instance, if the robot were to move the kettle instead of the cup, it performs the same action --- moving --- and would receive a high reward even though it actually fails to complete the task.}. Overall, \textbf{our} VIEW method was effective across the cluttered settings, grasping and manipulating the correct object to match the human's video demonstration in almost $100 \%$ of the trials. VIEW solved each task in less than $30$ minutes.

\begin{figure}
    \centering
    \includegraphics[width=0.98\columnwidth]{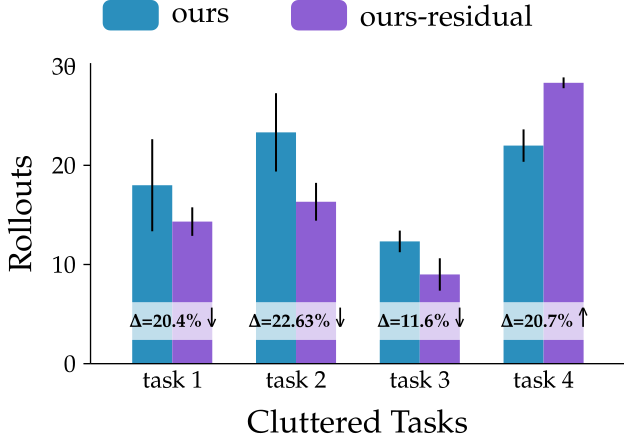}
    \caption{Experiment results for learning from multiple tasks. Here the robot has previously solved the \textit{Uncluttered} tasks, and it is now trying to learn a new \textit{Cluttered} task. We compare our VIEW algorithm without the residual (\textit{ours}) to our full VIEW algorithm with the residual (\textit{ours-residual}). There are a total of four video demonstration for different \textit{Cluttered} tasks. We plot the average number of rollouts needed for VIEW to solve each of these tasks. $\Delta$ is the percentage change in the number of rollouts with and without the residual. The bars indicate standard error of the mean.}
    \label{fig:exp_residual}
\end{figure}

\p{Learning from Multiple Tasks}
In \fig{exp_residual} we summarize the results from our final experiment. This experiment focused on how VIEW can leverage the tasks it has previously solved to improve its prior and accelerate its learning on new tasks. To quantify this acceleration, we measured the number of rollouts it took for the robot to successfully imitate a video demonstration in the \textit{Cluttered} environment. Both \textbf{ours} and \textbf{ours-residual} used VIEW, but \textbf{ours-residual} included the full VIEW algorithm with the residual component. We found that for $3$ out of the $4$ \textit{Cluttered} demonstrations, applying the residual significantly reduced the number of interactions needed to learn the task (roughly $25 \%$ fewer rollouts). Interestingly, for the fourth demonstration the residual actually had the opposite effect, and slowed down the robot's learning. On further examination, we believe this decrease in performance occurred because the initial hand trajectory $\xi^h$ lied outside the distribution of the data used to train the residual. Since the residual had not seen a demonstration that operated in the same part of the workspace, it was not able to de-noise the prior and accelerate the robot's learning. This suggests that --- while the residual can be useful --- it should be carefully applied. Learning a robust residual necessitates an expansive dataset that includes waypoints spanning the workspace.

\section{Limitations and Future Work} 

Our method has demonstrated the capability to expedite the learning process from human demonstrations, significantly reducing the required time from several hours \cite{whirl} to less than $30$ minutes. However, these results are achieved with certain limitations that present opportunities for future enhancements.

One primary limitation of our approach is its inability to incorporate orientation within the extracted trajectories. This restricts VIEW from handling tasks where orientation is crucial, such as pouring. The limitation stems from our discretization process for grasp exploration, which uses Centroidal Voronoi Tessellations to identify potential grasp points. This process does not extend to non-affine elements, such as orientation. Addressing this limitation in future work could involve incorporating Euler-Rodrigues formulas \cite{dai2015euler} to handle orientation changes in an affine space, allowing VIEW to extend to tasks requiring both positional and rotational considerations. By integrating orientation, the system could more effectively perform complex manipulation tasks.

Another limitation is VIEW’s dependence on a fixed camera pose between the human demonstration and the robot’s rollouts. Since our reward computation relies on object detection and pixel-based comparisons of object centroid locations, consistency in camera setup is essential to ensure accurate alignment between human demonstrations and robot actions. A promising direction to address this limitation is to integrate an object detector capable of pose estimation, such as PoseCNN \cite{cheron2015p} or Bundle-SDF \cite{wen2023bundlesdf}, which would allow for pose-based rather than purely pixel-based comparisons. By transforming object poses into a world-coordinate system that is invariant to camera angles, VIEW could adapt to scenarios with variable camera positions, further broadening its applicability.

Lastly, VIEW relies on environmental consistency between the demonstration and learning environments, which constrains the approach to task-specific setups. For example, if a demonstration shows a human picking up a cup from a specific location, the robot learns to perform the action from that exact position; any change in object location requires a new demonstration. To overcome this, future work could consider defining waypoints relative to object positions rather than in fixed $3$D coordinates. Combined with a pose-based reward function, this would enable the system to perform tasks even if object positions change.

% \textcolor{blue}{Despite these limitations, our method presents a promising step towards one-shot visual imitation learning. Our method can serve as a primer for downstream policy learning. Specifically, our method can quickly improve noisy priors using our exploration scheme, and train our residual network to model the noise distribution. A robust residual network can be used to extract state-action pairs directly from the video demonstration without the need for exploration. These state-action pairs can be used for behavior cloning or a policy learning framework to learn more adaptable policies. This integration would enable VIEW to data-efficiently translate human demonstrations into imitation learning policies that can respond to changes in the world state, serving as a foundational layer for broader task generalization.}

Despite these limitations, our method presents a promising step toward one-shot visual imitation learning. In addition, our method can also serve as a translation layer for downstream policy learning. Recall that a key limitation of policy learning approaches is the requirement for state-action pairs. VIEW can quickly generate these state-action pairs from video demonstrations, and with a robust residual network, we believe it can do so without the need for exploration. These state-action pairs could then be used in behavior cloning or a policy learning frameworks to learn more adaptable policies. This integration would enable VIEW to data-efficiently translate human demonstrations into imitation learning policies that can respond to changes in the world state, serving as a foundational layer for broader task generalization.

\section{Conclusion}

State-of-the-art visual imitation learning methods rely on intricate architectures to manage the complexities present in video demonstrations. This paper introduces an alternative framework designed to streamline the learning process by compressing video data and honing in on crucial features and waypoints. We show that by concentrating on these essential aspects, robots can more rapidly learn tasks from human video demonstrations. Our method, VIEW, incorporates distinct modules for (a) generating a condensed prior that captures the key aspects of the human demonstrator's intent, (b) facilitating targeted exploration around the waypoints in the prior through a division into grasp and task execution phases, and (c) employing a residual model to enhance learning efficiency by drawing on insights from previously completed tasks.

Through an ablation study in a simulated environment, we examine the contribution of each module to VIEW's overall efficacy. 
Our method achieves over $80 \%$ success rate when using our proposed trajectory compression, and it achieves over $90 \%$ success rate with our exploration approach. Further, using our residual network leads to a dramatic decrease in the number of rollouts (more than $40 \%$) needed to solve each task.
Subsequent real-world experiments, utilizing videos of human demonstrations, further validate our method's capability to effectively learn from such demonstrations.
In particular, our method achieves over $80 \%$ success rate in \textit{Uncluttered} tasks and achieves $100 \%$ success rate in \textit{Cluttered} tasks, outperforming the state-of-the-art baseline.
The combined results from our simulation studies and real-world testing indicate that VIEW can efficiently learn tasks demonstrated using a single video, typically requiring under $30$ minutes and fewer than $20$ real-world trials. Additionally, we advance the capabilities of human-to-robot visual imitation learning by showing that VIEW can learn from arbitrarily long video demonstrations involving multiple object interactions. These findings are illustrated in our supplemental videos, available here: \url{https://collab.me.vt.edu/view/}
\section{Declarations}

\p{Funding} This research was supported in part by the USDA National Institute of Food and Agriculture, Grant 2022-67021-37868.

\p{Conflict of Interest} The authors declare that they have no conflicts of interest.

\p{Ethical Statement} All physical experiments that relied on interactions with humans were conducted under university guidelines and followed the protocol of Virginia Tech IRB $\#20$-$755$.

\p{Author Contribution} A.J. led the algorithm development for prior extraction and agent agnostic reward computation. S.P. led the development for exploration. A.J. and S.P. wrote the first manuscript draft. A.J. ran the simulations and S.P. conducted the physical experiments. D.L. supervised the project, helped develop the method, and edited the manuscript.

\p{Acknowledgements} We thank Heramb Nemlekar for his valuable feedback on our manuscript.

\bibliographystyle{spmpsci}
\bibliography{citations}

\begin{thebibliography}{10}
\providecommand{\url}[1]{{#1}}
\providecommand{\urlprefix}{URL }
\expandafter\ifx\csname urlstyle\endcsname\relax
  \providecommand{\doi}[1]{DOI~\discretionary{}{}{}#1}\else
  \providecommand{\doi}{DOI~\discretionary{}{}{}\begingroup \urlstyle{rm}\Url}\fi

\bibitem{alakuijala2023learning}
Alakuijala, M., Dulac-Arnold, G., Mairal, J., Ponce, J., Schmid, C.: Learning reward functions for robotic manipulation by observing humans.
\newblock In: IEEE International Conference on Robotics and Automation, pp. 5006--5012 (2023)

\bibitem{amiranashvili2020scaling}
Amiranashvili, A., Dorka, N., Burgard, W., Koltun, V., Brox, T.: Scaling imitation learning in minecraft.
\newblock arXiv preprint arXiv:2007.02701  (2020)

\bibitem{whirl}
Bahl, S., Gupta, A., Pathak, D.: Human-to-robot imitation in the wild.
\newblock In: Robotics: Science and Systems (2022)

\bibitem{brown2019extrapolating}
Brown, D., Goo, W., Nagarajan, P., Niekum, S.: Extrapolating beyond suboptimal demonstrations via inverse reinforcement learning from observations.
\newblock In: International Conference on Machine Learning, pp. 783--792 (2019)

\bibitem{brown2020better}
Brown, D.S., Goo, W., Niekum, S.: Better-than-demonstrator imitation learning via automatically-ranked demonstrations.
\newblock In: Conference on Robot Learning, pp. 330--359 (2020)

\bibitem{caba2015activitynet}
Caba~Heilbron, F., Escorcia, V., Ghanem, B., Carlos~Niebles, J.: Activitynet: {A} large-scale video benchmark for human activity understanding.
\newblock In: IEEE Conference on Computer Vision and Pattern Recognition, pp. 961--970 (2015)

\bibitem{ycb}
Calli, B., Singh, A., Bruce, J., Walsman, A., Konolige, K., Srinivasa, S., Abbeel, P., Dollar, A.M.: {Yale-CMU-B}erkeley dataset for robotic manipulation research.
\newblock The International Journal of Robotics Research \textbf{36}(3), 261--268 (2017)

\bibitem{cetin2021domain}
Cetin, E., Celiktutan, O.: Domain-robust visual imitation learning with mutual information constraints.
\newblock In: International Conference on Learning Representations (2021)

\bibitem{chane2023learning}
Chane-Sane, E., Schmid, C., Laptev, I.: Learning video-conditioned policies for unseen manipulation tasks.
\newblock In: International Conference on Robotics and Automation, pp. 909--916 (2023)

\bibitem{chen2019deep}
Chen, J., Yuan, B., Tomizuka, M.: {D}eep imitation learning for autonomous driving in generic urban scenarios with enhanced safety.
\newblock In: IEEE/RSJ International Conference on Intelligent Robots and Systems, pp. 2884--2890 (2019)

\bibitem{cheron2015p}
Ch{\'e}ron, G., Laptev, I., Schmid, C.: P-cnn: Pose-based cnn features for action recognition.
\newblock In: IEEE International Conference on Computer Vision, pp. 3218--3226 (2015)

\bibitem{dai2015euler}
Dai, J.S.: Euler--rodrigues formula variations, quaternion conjugation and intrinsic connections.
\newblock Mechanism and Machine Theory \textbf{92}, 144--152 (2015)

\bibitem{das2013thousand}
Das, P., Xu, C., Doell, R.F., Corso, J.J.: A thousand frames in just a few words: {L}ingual description of videos through latent topics and sparse object stitching.
\newblock In: IEEE Conference on Computer Vision and Pattern Recognition, pp. 2634--2641 (2013)

\bibitem{duan2023ar2}
Duan, J., Wang, Y.R., Shridhar, M., Fox, D., Krishna, R.: Ar2-d2: {T}raining a robot without a robot.
\newblock arXiv preprint arXiv:2306.13818  (2023)

\bibitem{dulac2021challenges}
Dulac-Arnold, G., Levine, N., Mankowitz, D.J., Li, J., Paduraru, C., Gowal, S., Hester, T.: Challenges of real-world reinforcement learning: {d}efinitions, benchmarks and analysis.
\newblock Machine Learning \textbf{110}(9), 2419--2468 (2021)

\bibitem{fang2019survey}
Fang, B., Jia, S., Guo, D., Xu, M., Wen, S., Sun, F.: {S}urvey of imitation learning for robotic manipulation.
\newblock International Journal of Intelligent Robotics and Applications \textbf{3}, 362--369 (2019)

\bibitem{fontaine2020covariance}
Fontaine, M.C., Togelius, J., Nikolaidis, S., Hoover, A.K.: Covariance matrix adaptation for the rapid illumination of behavior space.
\newblock In: Genetic and Evolutionary Computation Conference, pp. 94--102 (2020)

\bibitem{gouda2022dopose}
Gouda, A., Ghanem, A., Reining, C.: {DoPose-6D} dataset for object segmentation and 6{D} pose estimation.
\newblock In: IEEE International Conference on Machine Learning and Applications, pp. 477--483 (2022)

\bibitem{gouda2023dounseen}
Gouda, A., Roidl, M.: Dounseen: {Z}ero-shot object detection for robotic grasping.
\newblock arXiv preprint arXiv:2304.02833  (2023)

\bibitem{goyal2017something}
Goyal, R., Ebrahimi~Kahou, S., Michalski, V., Materzynska, J., Westphal, S., Kim, H., Haenel, V., Fruend, I., Yianilos, P., Mueller-Freitag, M.: The" something something" video database for learning and evaluating visual common sense.
\newblock In: IEEE International Conference on Computer Vision, pp. 5842--5850 (2017)

\bibitem{habibian2022here}
Habibian, S., Jonnavittula, A., Losey, D.P.: Here’s what i’ve learned: Asking questions that reveal reward learning.
\newblock ACM Transactions on Human-Robot Interaction (THRI) \textbf{11}(4), 1--28 (2022)

\bibitem{he2017mask}
He, K., Gkioxari, G., Doll{\'a}r, P., Girshick, R.: Mask {R-CNN}.
\newblock In: IEEE international Conference on Computer Vision, pp. 2961--2969 (2017)

\bibitem{hussein2017imitation}
Hussein, A., Gaber, M.M., Elyan, E., Jayne, C.: {I}mitation learning: {A} survey of learning methods.
\newblock ACM Computing Surveys \textbf{50}(2), 1--35 (2017)

\bibitem{isola2017image}
Isola, P., Zhu, J.Y., Zhou, T., Efros, A.A.: Image-to-image translation with conditional adversarial networks.
\newblock In: IEEE Conference on Computer Vision and Pattern Recognition, pp. 1125--1134 (2017)

\bibitem{jain2024vid2robot}
Jain, V., Attarian, M., Joshi, N.J., Wahid, A., Driess, D., Vuong, Q., Sanketi, P.R., Sermanet, P., Welker, S., Chan, C., et~al.: Vid2robot: {E}nd-to-end video-conditioned policy learning with cross-attention transformers.
\newblock arXiv preprint arXiv:2403.12943  (2024)

\bibitem{jin2020geometric}
Jin, J., Petrich, L., Dehghan, M., Jagersand, M.: A geometric perspective on visual imitation learning.
\newblock In: IEEE/RSJ International Conference on Intelligent Robots and Systems, pp. 5194--5200 (2020)

\bibitem{jonnavittula2021know}
Jonnavittula, A., Losey, D.P.: {I} know what you meant: {L}earning human objectives by (under) estimating their choice set.
\newblock In: IEEE International Conference on Robotics and Automation, pp. 2747--2753 (2021)

\bibitem{jonnavittula2021learning}
Jonnavittula, A., Losey, D.P.: {L}earning to share autonomy across repeated interaction.
\newblock In: IEEE/RSJ International Conference on Intelligent Robots and Systems, pp. 1851--1858 (2021)

\bibitem{jonnavittula2022sari}
Jonnavittula, A., Mehta, S.A., Losey, D.P.: {SARI: S}hared autonomy across repeated interaction.
\newblock ACM Transactions on Human-Robot Interaction \textbf{13}(2), 1--36 (2024)

\bibitem{kelly2019hg}
Kelly, M., Sidrane, C., Driggs-Campbell, K., Kochenderfer, M.J.: {HG-DA}gger: {I}nteractive imitation learning with human experts.
\newblock In: IEEE International Conference on Robotics and Automation, pp. 8077--8083 (2019)

\bibitem{kim2023giving}
Kim, M.J., Wu, J., Finn, C.: Giving robots a hand: {L}earning generalizable manipulation with eye-in-hand human video demonstrations.
\newblock arXiv preprint arXiv:2307.05959  (2023)

\bibitem{kober2013reinforcement}
Kober, J., Bagnell, J.A., Peters, J.: Reinforcement learning in robotics: A survey.
\newblock The International Journal of Robotics Research \textbf{32}(11), 1238--1274 (2013)

\bibitem{lee2022learning}
Lee, R., Abou-Chakra, J., Zhang, F., Corke, P.: Learning fabric manipulation in the real world with human videos.
\newblock arXiv preprint arXiv:2211.02832  (2022)

\bibitem{lee2019copy}
Lee, S., Oh, S.W., Won, D., Kim, S.J.: Copy-and-paste networks for deep video inpainting.
\newblock In: IEEE/CVF International Conference on Computer Vision, pp. 4413--4421 (2019)

\bibitem{li2021meta}
Li, J., Lu, T., Cao, X., Cai, Y., Wang, S.: Meta-imitation learning by watching video demonstrations.
\newblock In: International Conference on Learning Representations (2021)

\bibitem{liu2024ok}
Liu, P., Orru, Y., Paxton, C., Shafiullah, N.M.M., Pinto, L.: Ok-robot: {W}hat really matters in integrating open-knowledge models for robotics.
\newblock arXiv preprint arXiv:2401.12202  (2024)

\bibitem{liu2018imitation}
Liu, Y., Gupta, A., Abbeel, P., Levine, S.: Imitation from observation: {L}earning to imitate behaviors from raw video via context translation.
\newblock In: IEEE International Conference on Robotics and Automation, pp. 1118--1125 (2018)

\bibitem{lynch2020language}
Lynch, C., Sermanet, P.: Language conditioned imitation learning over unstructured data.
\newblock In: Robotics: Science and Systems (2020)

\bibitem{mehta2024waypoint}
Mehta, S.A., Habibian, S., Losey, D.P.: Waypoint-based reinforcement learning for robot manipulation tasks.
\newblock arXiv preprint arXiv:2403.13281  (2024)

\bibitem{mehta2023unified}
Mehta, S.A., Losey, D.P.: Unified learning from demonstrations, corrections, and preferences during physical human-robot interaction.
\newblock ACM Transactions on Human-Robot Interaction \textbf{13}(3), 1--25 (2023)

\bibitem{menda2019ensembledagger}
Menda, K., Driggs-Campbell, K., Kochenderfer, M.J.: Ensemble{DA}gger: {A} bayesian approach to safe imitation learning.
\newblock In: IEEE/RSJ International Conference on Intelligent Robots and Systems, pp. 5041--5048 (2019)

\bibitem{monfort2021multi}
Monfort, M., Pan, B., Ramakrishnan, K., Andonian, A., McNamara, B.A., Lascelles, A., Fan, Q., Gutfreund, D., Feris, R.S., Oliva, A.: Multi-moments in time: Learning and interpreting models for multi-action video understanding.
\newblock IEEE Transactions on Pattern Analysis and Machine Intelligence \textbf{44}(12), 9434--9445 (2021)

\bibitem{morales2021survey}
Morales, E.F., Murrieta-Cid, R., Becerra, I., Esquivel-Basaldua, M.A.: A survey on deep learning and deep reinforcement learning in robotics with a tutorial on deep reinforcement learning.
\newblock Intelligent Service Robotics \textbf{14}(5), 773--805 (2021)

\bibitem{muckell2014compression}
Muckell, J., Olsen, P.W., Hwang, J.H., Lawson, C.T., Ravi, S.: Compression of trajectory data: {A} comprehensive evaluation and new approach.
\newblock GeoInformatica \textbf{18}, 435--460 (2014)

\bibitem{padalkar2023open}
Padalkar, A., Pooley, A., Jain, A., Bewley, A., Herzog, A., Irpan, A., Khazatsky, A., Rai, A., Singh, A., Brohan, A., et~al.: Open x-embodiment: {R}obotic learning datasets and rt-x models.
\newblock arXiv preprint arXiv:2310.08864  (2023)

\bibitem{pan2020imitation}
Pan, Y., Cheng, C.A., Saigol, K., Lee, K., Yan, X., Theodorou, E.A., Boots, B.: {I}mitation learning for agile autonomous driving.
\newblock The International Journal of Robotics Research \textbf{39}(2-3), 286--302 (2020)

\bibitem{pari2021surprising}
Pari, J., Shafiullah, N.M., Arunachalam, S.P., Pinto, L.: The surprising effectiveness of representation learning for visual imitation.
\newblock In: Robotics: Science and Systems (2021)

\bibitem{Patel2022}
Patel, A., Wang, A., Radosavovic, I., Malik, J.: Learning to imitate object interactions from internet videos.
\newblock arXiv preprint arXiv:2211.13225  (2022)

\bibitem{pomerleau1991efficient}
Pomerleau, D.A.: {E}fficient training of artificial neural networks for autonomous navigation.
\newblock Neural Computation \textbf{3}(1), 88--97 (1991)

\bibitem{rafailov2021visual}
Rafailov, R., Yu, T., Rajeswaran, A., Finn, C.: Visual adversarial imitation learning using variational models.
\newblock Advances in Neural Information Processing Systems \textbf{34}, 3016--3028 (2021)

\bibitem{ratliff2007imitation}
Ratliff, N., Bagnell, J.A., Srinivasa, S.S.: {I}mitation learning for locomotion and manipulation.
\newblock In: IEEE-RAS International Conference on Humanoid Robots, pp. 392--397 (2007)

\bibitem{ren2015faster}
Ren, S., He, K., Girshick, R., Sun, J.: Faster {R-CNN}: Towards real-time object detection with region proposal networks.
\newblock IEEE transactions on pattern analysis and machine intelligence \textbf{39}(6), 1137--1149 (2016)

\bibitem{romero2022embodied}
Romero, J., Tzionas, D., Black, M.J.: Embodied hands: {M}odeling and capturing hands and bodies together.
\newblock ACM Transactions on Graphics \textbf{36}(6) (2017)

\bibitem{rong2021frankmocap}
Rong, Y., Shiratori, T., Joo, H.: Frankmocap: {A} monocular 3d whole-body pose estimation system via regression and integration.
\newblock In: IEEE International Conference on Computer Vision Workshops, pp. 1749--1759 (2021)

\bibitem{ross2011reduction}
Ross, S., Gordon, G., Bagnell, D.: A reduction of imitation learning and structured prediction to no-regret online learning.
\newblock In: International Conference on Artificial Intelligence and Statistics, pp. 627--635 (2011)

\bibitem{schaal1996learning}
Schaal, S.: {L}earning from demonstration.
\newblock In: Advances in Neural Information Processing Systems, vol.~9 (1996)

\bibitem{schafer2023visual}
Sch{\"a}fer, L., Jones, L., Kanervisto, A., Cao, Y., Rashid, T., Georgescu, R., Bignell, D., Sen, S., Gavito, A.T., Devlin, S.: Visual encoders for data-efficient imitation learning in modern video games.
\newblock arXiv preprint arXiv:2312.02312  (2023)

\bibitem{scheller2020sample}
Scheller, C., Schraner, Y., Vogel, M.: Sample efficient reinforcement learning through learning from demonstrations in minecraft.
\newblock In: NeurIPS Competition and Demonstration Track, pp. 67--76 (2020)

\bibitem{sermanet2017unsupervised}
Sermanet, P., Xu, K., Levine, S.: Unsupervised perceptual rewards for imitation learning.
\newblock In: Robotics: Science and Systems (2017)

\bibitem{shafiullah2023bringing}
Shafiullah, N.M.M., Rai, A., Etukuru, H., Liu, Y., Misra, I., Chintala, S., Pinto, L.: On bringing robots home.
\newblock arXiv preprint arXiv:2311.16098  (2023)

\bibitem{shan2020understanding}
Shan, D., Geng, J., Shu, M., Fouhey, D.F.: Understanding human hands in contact at internet scale.
\newblock In: IEEE/CVF Conference on Computer Vision and Pattern Recognition, pp. 9869--9878 (2020)

\bibitem{sharma2019third}
Sharma, P., Pathak, D., Gupta, A.: Third-person visual imitation learning via decoupled hierarchical controller.
\newblock In: Advances in Neural Information Processing Systems, vol.~32 (2019)

\bibitem{shaw2024learning}
Shaw, K., Bahl, S., Sivakumar, A., Kannan, A., Pathak, D.: Learning dexterity from human hand motion in internet videos.
\newblock The International Journal of Robotics Research \textbf{43}(4), 513--532 (2024)

\bibitem{shi2024yell}
Shi, L.X., Hu, Z., Zhao, T.Z., Sharma, A., Pertsch, K., Luo, J., Levine, S., Finn, C.: Yell at your robot: {I}mproving on-the-fly from language corrections.
\newblock arXiv preprint arXiv:2403.12910  (2024)

\bibitem{shi2023waypoint}
Shi, L.X., Sharma, A., Zhao, T.Z., Finn, C.: Waypoint-based imitation learning for robotic manipulation.
\newblock In: Conference on Robot Learning (2023)

\bibitem{sieb2020graph}
Sieb, M., Xian, Z., Huang, A., Kroemer, O., Fragkiadaki, K.: Graph-structured visual imitation.
\newblock In: Conference on Robot Learning, pp. 979--989 (2020)

\bibitem{smith2019avid}
Smith, L., Dhawan, N., Zhang, M., Abbeel, P., Levine, S.: {AVID}: {L}earning multi-stage tasks via pixel-level translation of human videos.
\newblock In: Robotics: Science and Systems (2020)

\bibitem{BO}
Snoek, J., Larochelle, H., Adams, R.P.: Practical bayesian optimization of machine learning algorithms.
\newblock In: Advances in Neural Information Processing Systems, vol.~25 (2012)

\bibitem{song2020grasping}
Song, S., Zeng, A., Lee, J., Funkhouser, T.: Grasping in the wild: {L}earning 6dof closed-loop grasping from low-cost demonstrations.
\newblock IEEE Robotics and Automation Letters \textbf{5}(3), 4978--4985 (2020)

\bibitem{taranovic2022adversarial}
Taranovic, A., Kupcsik, A.G., Freymuth, N., Neumann, G.: Adversarial imitation learning with preferences.
\newblock In: International Conference on Learning Representations (2022)

\bibitem{tremblay2018falling}
Tremblay, J., To, T., Birchfield, S.: Falling things: A synthetic dataset for 3d object detection and pose estimation.
\newblock In: IEEE Conference on Computer Vision and Pattern Recognition Workshops, pp. 2038--2041 (2018)

\bibitem{vassiliades2017using}
Vassiliades, V., Chatzilygeroudis, K., Mouret, J.B.: Using centroidal voronoi tessellations to scale up the multidimensional archive of phenotypic elites algorithm.
\newblock IEEE Transactions on Evolutionary Computation \textbf{22}(4), 623--630 (2017)

\bibitem{wang2020rgb2hands}
Wang, J., Mueller, F., Bernard, F., Sorli, S., Sotnychenko, O., Qian, N., Otaduy, M.A., Casas, D., Theobalt, C.: Rgb2hands: {R}eal-time tracking of 3d hand interactions from monocular rgb video.
\newblock ACM Transactions on Graphics \textbf{39}(6), 1--16 (2020)

\bibitem{wen2022you}
Wen, B., Lian, W., Bekris, K., Schaal, S.: You only demonstrate once: {C}ategory-level manipulation from single visual demonstration.
\newblock In: Robotics: Science and Systems (2022)

\bibitem{wen2023bundlesdf}
Wen, B., Tremblay, J., Blukis, V., Tyree, S., M{\"u}ller, T., Evans, A., Fox, D., Kautz, J., Birchfield, S.: Bundlesdf: Neural 6-dof tracking and 3d reconstruction of unknown objects.
\newblock In: IEEE/CVF Conference on Computer Vision and Pattern Recognition, pp. 606--617 (2023)

\bibitem{wen2021keyframe}
Wen, C., Lin, J., Qian, J., Gao, Y., Jayaraman, D.: Keyframe-focused visual imitation learning.
\newblock In: International Conference on Machine Learning, vol. 139, pp. 11123--11133 (2021)

\bibitem{xiong2021learning}
Xiong, H., Li, Q., Chen, Y.C., Bharadhwaj, H., Sinha, S., Garg, A.: Learning by watching: {P}hysical imitation of manipulation skills from human videos.
\newblock In: IEEE/RSJ International Conference on Intelligent Robots and Systems, pp. 7827--7834 (2021)

\bibitem{young2021visual}
Young, S., Gandhi, D., Tulsiani, S., Gupta, A., Abbeel, P., Pinto, L.: Visual imitation made easy.
\newblock In: Conference on Robot Learning (2021)

\bibitem{zhang2020mediapipe}
Zhang, F., Bazarevsky, V., Vakunov, A., Tkachenka, A., Sung, G., Chang, C.L., Grundmann, M.: Mediapipe hands: {O}n-device real-time hand tracking.
\newblock In: CVPR Workshop on Computer Vision for Augmented and Virtual Reality (2020)

\end{thebibliography}

\appendix
\section{Appendix} \label{sec:appendix}

\begin{figure}[t]
    \centering
    \includegraphics[width=0.98\columnwidth]{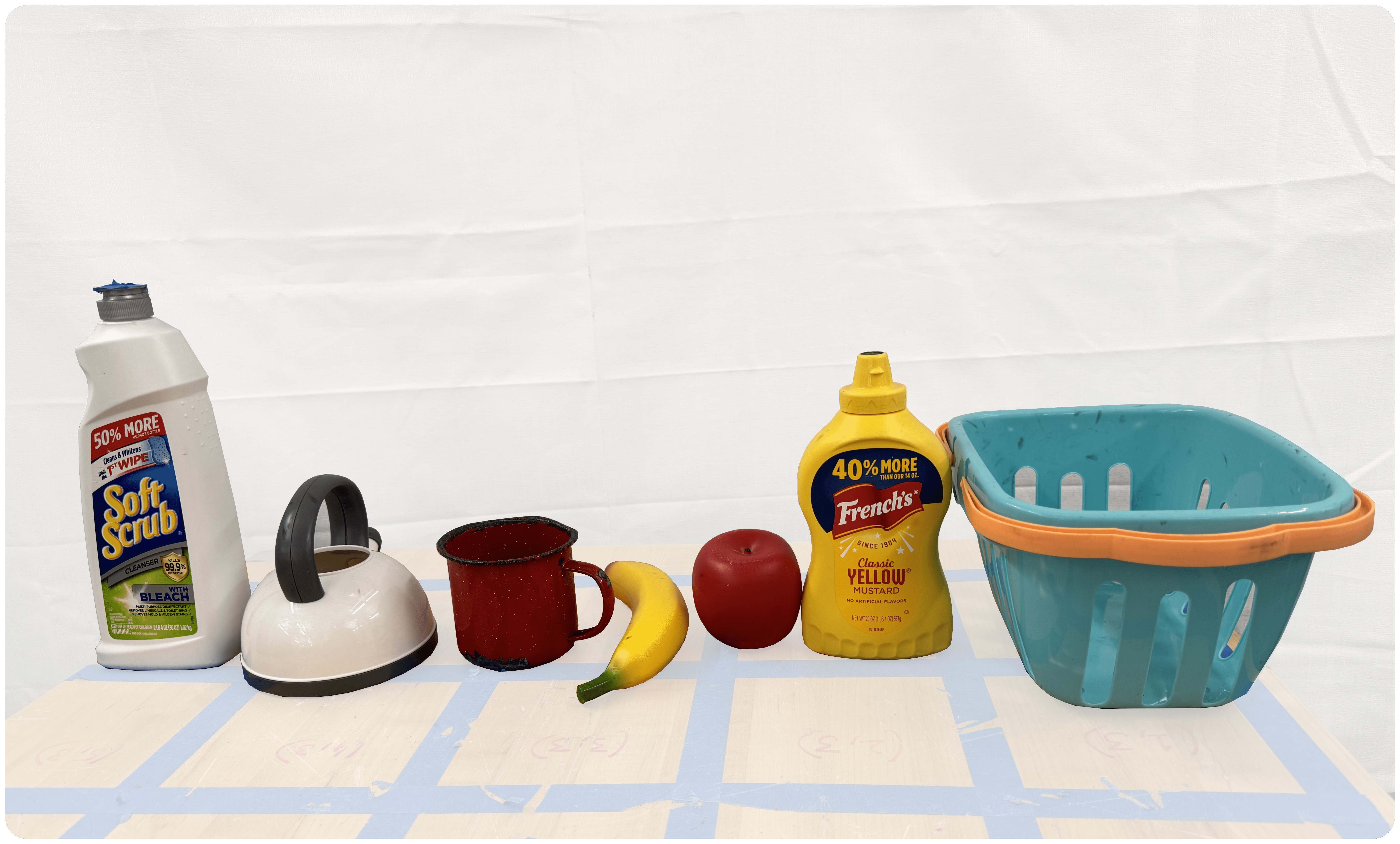}
    \caption{Objects manipulated in our real-world experiments. (From left to right) We use a bottle of bleach, a kettle, a mug, a banana, an apple, a bottle of mustard, and a basket. These seven distinct items were systematically selected for assessment based on their varying shapes, sizes, and colors to provide a comprehensive evaluation of our algorithm.}
    \label{fig:app_objects}
\end{figure}

\begin{figure*}
    \centering
    \includegraphics[width=2\columnwidth]{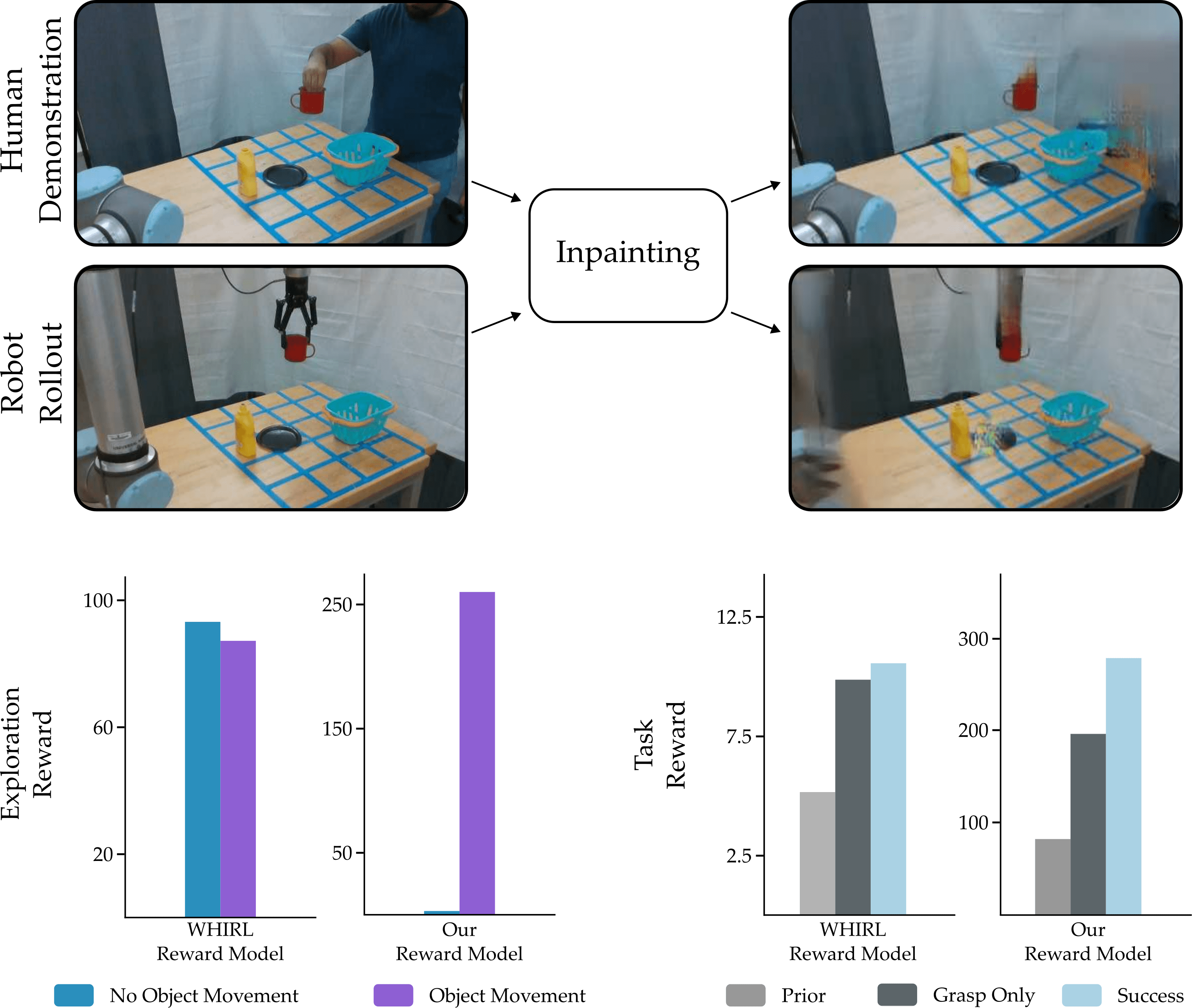}
    \caption{Challenges with WHIRL Evaluation. (Top) As described in \cite{whirl}, we utilized Copy-Paste Networks \cite{lee2019copy} for the purpose of video inpainting, with the aim of removing both the human demonstrator and the robot arm from the video frames. This process is critical for enabling the comparison of frames through moment models \cite{monfort2021multi}, which in turn facilitates the computation of agent-agnostic rewards. However, in our evaluations we encountered consistency issues with the inpainted images, leading to highly variable reward signals. (Bottom Left) The inconsistency in reward signals led to scenarios where the robot received high exploration rewards without actually moving the object. This is problematic because the robot relies on these rewards to identify waypoints that are near the object, which are necessary for successful grasping. In contrast, WHIRL with our reward model produces low exploration rewards when there is no object movement, and rewards increase significantly only when the object is displaced. This variability in the WHIRL reward model often caused the robot's learning trajectory to converge prematurely at a suboptimal point, usually far from the target object. (Bottom Right) When the robot managed to overcome the variability in exploration rewards and successfully grasped the object, we observed that the reward difference between just grasping the object and completing the entire task was minimal. WHIRL with our reward model provided a clearer distinction between these different phases of the task. The lack of clear reward differentiation in WHIRL's reward model frequently hindered the robot's ability to fully learn the task, often resulting in the robot only learning to pick up the object without completing subsequent steps. Based on these results, in our experiments from Section~\ref{sec:experiments} we used WHIRL with our proposed reward model instead of WHIRL with its original reward model.}
    \label{fig:app_whirl}
\end{figure*}

\subsection{Implementation Details}

A public repository of our code can be found here: \\ \url{https://github.com/VT-Collab/view}

\p{Data collection} In our experiments, we use the Intel RealSense D435 RGB-D camera to capture video demonstrations, recording both RGB and depth data at 60 frames per second for each demonstration and rollout. While the depth data can be noisy, the Intel RealSense SDK offers several filters to enhance quality; we found that the hole-filling filter provided the most significant improvement. Despite this filtering, noise still introduced deviations of 5 to 19 cm from the ground truth waypoints. All our experiments incorporated this level of noise in the extracted waypoints, and our method, VIEW, effectively denoised these waypoints to achieve reliable results.

\p{Hand trajectory extraction} In line with the methodology described in WHIRL \cite{whirl}, we utilize the 100 Days of Hands (100DOH) detector from \url{https://github.com/ddshan/hand_detector.d2} for identifying hand-object contact points. For wrist detection, we integrate this with FrankMocap, as documented in Rong \textit{et al.} \cite{rong2021frankmocap}, without any model fine-tuning. The implementation for FrankMocap can be found here: \url{https://github.com/facebookresearch/frankmocap}. To obtain compressed trajectories, we combine the output of the FrankMocap model with SQUISHE. We develop our own version of SQUISHE based on the description provided in \cite{muckell2014compression}. Our implementation can be accessed in the code repository.

\p{Object trajectory extraction} To identify objects within the scene, we use Mask R-CNN, as detailed by He \textit{et al.} \cite{he2017mask}, through its implementation in Detectron2 (\url{https://github.com/facebookresearch/detectron2}). Following the methodologies outlined in \cite{gouda2023dounseen}, we initially pretrain our model using the Nvidia Falling Things dataset \cite{tremblay2018falling} and the DoPose-6D dataset \cite{gouda2022dopose}. We then finetune the model on a custom dataset containing $21$ objects, with a subset of $7$ being directly relevant to our final evaluations. This subset includes standard objects from the YCB object dataset \cite{ycb} and others that are commonly found in kitchen environments. The complete list of objects used in our evaluation is shown in \fig{app_objects}.

\p{Object detection during runtime} Given the challenges of ensuring object visibility and detection in every frame, we detect the object’s location only at key waypoints identified through compression with SQUISHE, interpolating for all other points along the trajectory. This approach allows us to perform object detection at select locations, significantly reducing the computational load during runtime.

\p{Residual network} For our residual network, we employ a fully connected multi-layer perceptron with two hidden layers, utilizing ReLU as the activation function and mean squared error (MSE) for loss calculation. We use the Adam optimizer and train the network for $100$ epochs. The initial learning rate is set at $0.1$, with a decay factor of $0.15$. Our network takes each waypoint as an input and then outputs the corrected waypoint. We use the same residual network across multiple tasks and multiple objects. For more detailed information on our training parameters, please refer to our code repository.

\p{Robot Experiments} In our robot experiments, during the grasp exploration the robot searches for a good grasp location. At the end of each rollout, proctors manually reset the objects to their original positions ensuring minimal errors in the object positions. Once the robot identifies a successful grasp location, the environment reset is automated, with the robot itself returning the item to its original position. The robot movement in these exploration rollouts are executed with a compliance controller to present any damages to the robot or the objects.

\subsection{Challenges with WHIRL} 

Because of the lack of publicly available implementations of WHIRL, we developed our version based on the algorithms provided in WHIRL's publication \cite{whirl}. As described, we used a four-layer MLP, implemented as a Variational Autoencoder and optimized via KL divergence loss. Initially --- consistent with the guidelines in WHIRL's manuscript --- we employed Copy-Paste Networks for inpainting \cite{lee2019copy} and the moment model from Monfort \textit{et al.} \cite{monfort2021multi} for calculating rewards. 

However, during our experiments, we encountered two major challenges with WHIRL (see \fig{app_whirl}). The first issue was the inconsistency observed in the video inpainting performance, where the Copy-Paste Network failed to fully remove the robot from several frames (See \fig{app_whirl} Top). This inconsistency persisted even after we fine-tuned the model on $400$ custom images of our robot. Having consistent images with the human and the robot removed are particularly critical because WHIRL's exploration strategy relies heavily on comparing embeddings across frames. Due to the erratic inpainting results, the robot often converged to suboptimal positions, distant from the target object (See \fig{app_whirl} Bottom Left). The second issue pertained to the rewards linked with task completion. There was a marked lack of differentiation in rewards between trajectories where the robot only grasped the object and those where it successfully completed the task (see \fig{app_whirl} Bottom Right). This similarity in rewards often caused the robot to become stuck in a local minimum, proficient at object pickup but failing to complete the rest of the manipulation task.

In response to these issues, we replaced the reward model from WHIRL with the reward model described in Section~\ref{sec:view_rewards}. While WHIRL calculates exploration rewards based on the variance in frame embeddings and task rewards through the difference in video embeddings, our method takes a different approach. We explicitly compute exploration rewards using changes in the object’s position and gauge task completion by measuring the object's proximity to the demonstrated trajectory. Our reward structure therefore capitalizes on direct and pertinent information (i.e., the location of the target object) rather than an indeterminate high-dimensional representation. We conducted a limited set of experiments to ensure that reward responses from our model were comparable to those from the original moment model. We believe that this modification does not fundamentally alter the functionality of WHIRL, and is a reasonable baseline for comparison.

\subsection{Algorithms}\label{app:algorithms}
\begin{algorithm}
\caption{VIEW}
\label{alg:view}
    \hspace*{\algorithmicindent} \textbf{Input:} Video of human interaction $V$ \\
    \hspace*{\algorithmicindent} Residual network $\Phi$ \\
    \hspace*{\algorithmicindent} Dataset of previous corrections $\mathcal{D}$ \\
    \hspace*{\algorithmicindent} \textbf{Output:} Successful robot trajectory $\xi^r$\\
    \hspace*{\algorithmicindent} Updated Residual network $\Phi$ \\
    \begin{algorithmic}[1]
        \State $\xi^h$ = \Call{ExtractHandTraj}{$V$}
        \State $\xi^o$, $tag$ = \Call{ExtractObjectTraj}{$V$, $\xi^h$}
        \State $\xi^h_c$ = $\xi^h$ + \Call{$\Phi$}{$\xi^h$}
        \State $\xi^h_{grasp}$, $\xi^h_{task}$ = \Call{DivideTraj}{$\xi^h_c$}
        \State $\xi^*_{grasp}$ = \Call{GraspExplore}{$\xi^h_{task}$, $\xi^o$, $tag$}
        \If{$\xi^*_{grasp}$ is success}
            \State $\xi^*_{task}$ = \Call{TaskExplore}{$\xi^h_{task}$, $\xi^*_{grasp}$, $\xi^o$, $tag$}
            \If{$\xi^*_{task}$ is success}
                \State $\xi^*$ = \Call{CombineTraj}{$\xi^*_{grasp}$, $\xi^*_{task}$}
                \State Add ($\xi^h$ , $\xi^*$) to $\mathcal{D}$
                \State Retrain $\Phi$ on $\mathcal{D}$
            \EndIf
        \EndIf
    \end{algorithmic}
\end{algorithm}

\begin{algorithm}
\caption{Object Trajectory Extraction}
\label{alg:obj_traj}
    \hspace*{\algorithmicindent} \textbf{Input:} Video of human interaction $V$ \\
    \hspace*{\algorithmicindent} Depth information $D^h$ \\
    \hspace*{\algorithmicindent} Human hand trajectory $\xi^h$ \\
    \hspace*{\algorithmicindent} Object detection model $OD$ \\
    \hspace*{\algorithmicindent} Set of Anchor boxes $A$ \\
    \hspace*{\algorithmicindent} Camera intrinsic and extrinsic parameters $\mathcal{C}^c_r$ \\
    \hspace*{\algorithmicindent} \textbf{Output:} Object $tag$\\
    \hspace*{\algorithmicindent} Object trajectory in pixel space $\zeta^o_h$\\
    \hspace*{\algorithmicindent}  Object trajectory in 3D space $\xi^o_h$\\
    \begin{algorithmic}[1]
        \State Initialize $OD$
        \State Initialize object count
        \For{contact information $c_t$ in $\xi^h$ if $c_t$ = True}
            \State $v_t$ = \Call{ExtractVideoFrame}{$V, c_t$}
            \For{Anchor box $\alpha$ in $A$}
                \State objects = \Call{OD}{$\alpha$}
                \State Update object count
            \EndFor
        \EndFor
        \State $tag$ = \Call{Max}{object count}
        \State $\xi^o_h$, $\zeta^o_h$ = \Call{ExtractObjectTraj}{$V$ ,$tag$}
        \State \Return $\xi^o_h$, $\zeta^o_h$, $tag$
        \State
        \Function{ExtractObjectTraj}{$V$, $tag$}
            \State Initialize an empty lists $\xi_h$, $\zeta_h$
            \For {Video frame $v_t$ in $V$}
                \State $p^o_{x_t}$, $p^o_{y_t}$ = \Call{$OD$}{$v_t, tag$}
                \State Append ($p^o_{x_t}$, $p^o_{y_t}$) to $\zeta_h$
                \State $\delta^o_t$ = \Call{$D^h$}{$p^o_{x_t}$, $p^o_{y_t}$}
                \State $x^o_t, y^o_t, z^o_t$ = \Call{$\mathcal{C}^c_r$} {$p^o_{x_t}$, $p^o_{y_t}$, $\delta^o_t$}
                \State Append ($x^o_t, y^o_t, z^o_t$) to $\xi^o_i$
            \EndFor
            \State \Return $\xi_h$, $\zeta_h$
        \EndFunction
    \end{algorithmic}
\end{algorithm}

\begin{algorithm}
\caption{Grasp Exploration}
\label{alg:grasp_exp}
    \hspace*{\algorithmicindent} \textbf{Input:} Prior trajectory of grasping $\xi^h_{grasp}$ \\
    \hspace*{\algorithmicindent} object location in robot coordinates $\omega^o$ \\
    \begin{algorithmic}[1]
        \State Initialize $\delta, \epsilon, p_{explore}$
        \State Initialize flag $local = False$
        \State Initialize an empty list $\Omega^r_{visited}$
        \State Get the point where human grasps the object $\omega^h_{grasp}$ from prior $\xi^h_{grasp}$
        % \State Calculate unit vector using \eq{unit_vec}
        \State Define bounding box $\mathcal{B}$ that circumscribes the points $\omega^o_{grasp} - \Delta\hat{j}$ and $\omega^h_{grasp} + \Delta\hat{j}$
        \State Sample random $points$ from $\mathcal{B}$ and initialize the set $\Omega^r_{unvisited} = \Call{K-Means}{points}$
        \State Initialize Bayesian optimizer \Call{BO}{}
        \State

        \Function{Ask}{}
            \State Generate $p$ from uniform distribution $[0, 1]$
            \If{$p < p_{explore}$}
                \State Sample high-level waypoint $\omega^r$ from $\Omega^r_{unvisited}$ using \eq{sampling_strategy}
                \State Change flag $local = False$
                \State \Return $\omega^r$
            \Else
                \State Sample high-level waypoint from $\Omega^r_{visited}$ with probability distribution from \eq{p_sample_visited}
                \State Start low-level search by defining a bounding box around this waypoint with limits $\epsilon$
                \State Query \Call{BO}{} for $\omega^r$
                \State Change flag $local = True$
                \State \Return $\omega^r$
            \EndIf
        \EndFunction
        \State
        
        \Function{Tell}{$\omega^r_i, R_i$}
            \If{$local$}
                \State Update \Call{BO}{} with $\omega^r_i, R_i$
            \Else{}
                \State Remove $\omega^r_i$ from $\Omega^r_{unvisited}$
                \State Add $(\omega^r_i, R_i)$ to $\Omega^r_{visited}$
            \EndIf
        \EndFunction
        \State

        \While{\textit{grasp} is not successful}
        \State Sample a waypoint $\omega^r = \Call{Ask}{}$
        \State Execute trajectory $\xi^r = $ 
        \State $\quad\quad \{ \omega^r_{t_1}, \omega^r_{t_2}, \dots \omega^r, \omega^r_{grasp + 1} \}$
        \State Get the reward $R$ using \eq{reward}
        \State Inform the explorer $\Call{Tell}{\omega^r, R}$

        \EndWhile
    \end{algorithmic}
\end{algorithm}

\begin{algorithm}
\caption{Task Exploration}
\label{alg:task_exp}
    \hspace*{\algorithmicindent} \textbf{Input:} Prior trajectory of task $\xi^h_{task}$
    \begin{algorithmic}[1]
        \State Define bounding box for each waypoint $\omega^r \in \xi^r_{task}$
        \State Initialize a separate Bayesian optimizer \Call{BO}{} for each waypoint in task
        \State

        \Function{Ask}{}
            \State Initialize an empty list $\xi^r_{task}$
            \For{$\omega^h_i \in \xi_{task}^h$}
                \State Query \Call{BO}{} for $\omega^r_i$
                \State Add $\omega^r_i$ to $\xi^r_{task}$
            \EndFor
            \State \Return $\xi^r_{task}$
        \EndFunction
        \State
        
        \Function{Tell}{$\xi^r_{task}, R$}
            \For{$i = 1, \dots, n$}
                \State Update corresponding \Call{BO}{} with $\omega^r_i \in \xi^r_{task}, R_i \in R$
            \EndFor
        \EndFunction
        \State

        \While{\textit{task} is not successful}
        \State Sample trajectory $\xi^r_{task} = \Call{Ask}{}$
        \State Execute trajectory $\xi^r_{task}$ in environment
        \State Get the reward $R$ for each waypoint in the trajectory using \eq{reward}
        \State Inform the explorer $\Call{Tell}{\xi^r_{task}, R}$

        \EndWhile
    \end{algorithmic}
\end{algorithm}


\end{document}